\documentclass[10pt,twocolumn,letterpaper]{article}

\usepackage{iccv}
\usepackage{times}
\usepackage{epsfig}
\usepackage{graphicx}
\usepackage{amsmath}
\usepackage{amssymb}
\usepackage{xspace}
\usepackage{microtype}

\usepackage[accsupp]{axessibility} %

\usepackage[pagebackref=true,breaklinks=true,letterpaper=true,colorlinks,bookmarks=false]{hyperref}

\iccvfinalcopy %

\def\iccvPaperID{2049} %

\ificcvfinal\fi

\newcommand{\ourdataset}{Doppelgangers\xspace}
\newcommand{\ourdatasetfull}{Doppelgangers dataset\xspace}
\newcommand{\twinterm}{doppelgangers\xspace}

\newenvironment{packed_item}{
\begin{itemize}
  \setlength{\itemsep}{1pt}
  \setlength{\parskip}{2pt}
  \setlength{\parsep}{0pt}
}{\end{itemize}}

\usepackage{pifont}
\usepackage{multirow}
\usepackage{multicol}
\usepackage{tabularx, booktabs}
\usepackage{enumitem}
\usepackage{float}

\begin{document}

\title{Doppelgangers: Learning to Disambiguate Images of Similar Structures}

\author{
Ruojin Cai$^1$ \quad
Joseph Tung$^1$  \quad
Qianqian Wang$^1$  \quad
\\[1mm]
Hadar Averbuch-Elor$^{2}$ \quad
Bharath Hariharan$^1$ \quad
Noah Snavely$^{1}$ 
\\[1mm]
$^1$Cornell University \ \ \
$^2$Tel Aviv University
}

\maketitle
\ificcvfinal\thispagestyle{empty}\fi

\begin{abstract}
    We consider the visual disambiguation task of determining whether a pair of visually similar images depict the same or distinct 3D surfaces (e.g., the same or opposite sides of a symmetric building). 
    Illusory image matches, where two images observe distinct but visually similar 3D surfaces, can be challenging for humans to differentiate, and can also lead 3D reconstruction algorithms to produce erroneous results.
    We propose a learning-based approach to visual disambiguation, formulating it as a binary classification task on image pairs.
    To that end, we introduce a new dataset for this problem, \ourdataset, which includes image pairs of similar structures with ground truth labels.
    We also design a network architecture that takes the spatial distribution of local keypoints and matches as input, allowing for better reasoning about both local and global cues.
    Our evaluation shows that our method can distinguish illusory matches in difficult cases, and can be integrated into SfM pipelines to produce correct, disambiguated 3D reconstructions.
    See our project page for our code, datasets, and more results: \href{http://doppelgangers-3d.github.io/}{doppelgangers-3d.github.io}.
\end{abstract}

\section{Introduction}

From time to time we are faced with the task of distinguishing between two things that are nearly indistinguishable, but are not the same object.
Examples of such lookalikes include identical twin siblings, two similar keys on a keychain, and two cups on a table at a party---only one of which is ours.
While these objects might look identical, there are often subtle cues we can use to tell them apart; for instance, even ``identical'' twins are not truly visually identical, but have perceptible differences.

\begin{figure}
    \centering
    \includegraphics[width=\columnwidth, trim=0 25 35 0, clip]{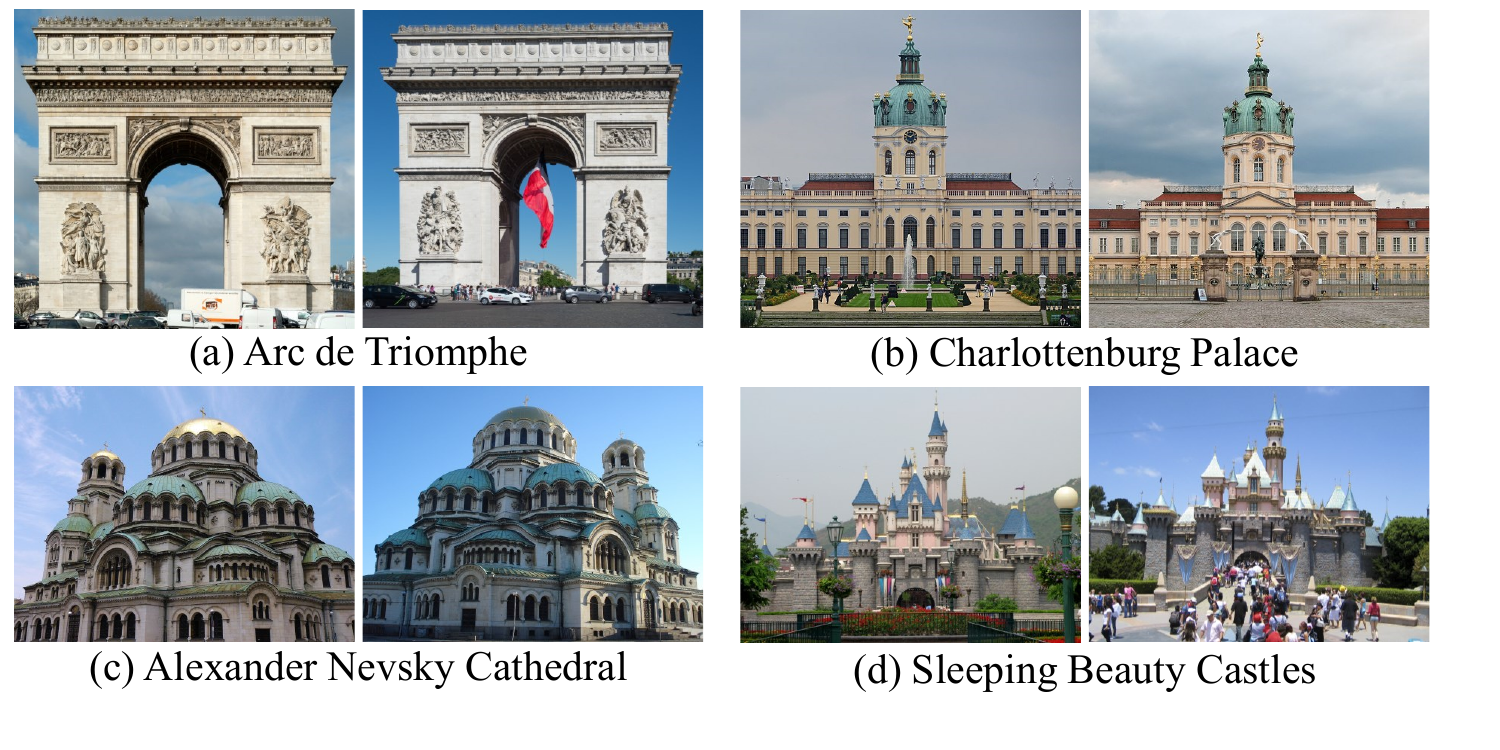}
    \vspace{-10pt}
   \caption[teaser]{
   \small
   These image pairs observe distinct but visually similar 3D surfaces.
   Can you spot the differences and distinguish between the two images in each pair?
   Hints in the footnote.\setcounter{footnote}{0}\footnotemark\ Such illusory image matches can fool humans, and also fool 3D reconstruction algorithms into thinking they share 3D correspondence.
   We propose a new method to disambiguate these kinds of false matches from image pairs that truly observe the same structure.
   }
\label{fig:teaser}
\end{figure}

Computer vision systems also face a version of this problem. In particular, our work considers geometric vision tasks like 3D reconstruction, where methods often must determine whether two images depict the exact same 3D surface in the world, or two different 3D surfaces that happen to look very similar---where wrong answers can lead to wrong 3D models. 
We call this task \emph{visual disambiguation}, but you could also call it the \emph{Big Ben problem}: London's Big Ben is a clock tower with four-way symmetry, where the four sides of the tower look nearly the same. 
Local feature matching methods like SIFT easily confuse one side for another, finding many matches between distinct 3D surfaces. 
These spurious matches lead structure from motion methods to produce incorrect reconstructions where multiple sides collapse together. Yet views of different sides of Big Ben are not truly identical---the individual bricks are different, the backgrounds are different, etc.
If matching methods knew what to look for, they could perhaps tell the different sides apart. 
Figure~\ref{fig:teaser} shows other examples of this ``spot the difference'' problem---see if you can tell these structures apart yourself.

\footnotetext{Attend to: (a) the sculptures on the bottom half of the monument, (b) the shape of the facade's roof (flat or triangular), (c) the location of the smaller tower (left or right), and (d) the background (with or without a mountain). Note that (a), (b), (c) show different faces of the same building, while (d) shows replicas of the same castle in different Disney parks.}

We call illusory image matches like those in Fig.~\ref{fig:teaser} \emph{\twinterm}, after the idea of two distinct people or objects that look very similar. Prior methods for 
disambiguating \twinterm in the context of 3D
vision have devised heuristics that analyze the structure of a full image collection. In contrast, our paper explores the fundamental building block of pairwise image comparison: can we automatically determine whether two views are the same, or just similar? We formulate this visual disambiguation problem as a binary classification task on image pairs, and develop a learning-based solution. 

Our solution involves assembling a new dataset, \emph{\ourdataset}, consisting of image pairs that either depict the same surface (positives) or two different, similar surfaces (negatives). 
Creating \ourdataset involved a challenging data curation task, since even humans can struggle to distinguish \emph{same} from \emph{similar}. We show how to use existing image annotations stored in the Wikimedia Commons image database to automatically create a large set of labeled image pairs.
We find that simply training a deep network model using these raw image pairs performs poorly. Therefore, 
we also design a network where we provide useful information 
in the form of local features and 2D correspondence.

On our \ourdataset test set, we find that our method works remarkably well on challenging disambiguation tasks, and significantly better than baselines and alternative network designs. 
We also explore the use of our learned classifier as a simple pre-processing filter on scene graphs computed in structure from motion pipelines like COLMAP~\cite{schonberger2016structure}, and find that it significantly improves the correctness of reconstructions on a set of difficult scenes, outperforming more complex visual disambiguation algorithms.

In summary, our paper makes the following contributions:
\begin{packed_item}
  \item We formulate the visual disambiguation problem on pairs of images.
  \item Based on this formulation, we create the \ourdataset Dataset, leveraging the existing cataloging of imagery on Wikimedia Commons.
  \item We design a network architecture well-suited for solving pairwise visual disambiguation as a classification problem. We show that training this network on our dataset leads to strong classification performance and downstream utility to SfM problems.
\end{packed_item}

\section{Related Work}\label{sec:related}

Local feature matching methods have been wildly successful~\cite{lowe2004distinctive}.
The combination of repeatability, discriminability, invariance to image transformations, and robustness to factors like partial occlusion make local features ideal for answering the question, \emph{Are these two images in correspondence?}, solidifying them as a foundation for downstream tasks like 
image retrieval~\cite{sivic2003videogoogle,nister2006scalable,chum2007totalrecall}, 
near-duplicate detection~\cite{chum2008near}, 
and structure from motion (SfM)~\cite{schaffalitzky2002multi,snavely2006photo,schonberger2016structure}.

However, these same properties make it hard for local feature matching methods to definitively answer the negation of this question: \emph{Are these two (possibly similar) images \textbf{not} in correspondence?} 
The dominant assumption in local feature matching is that sufficiently many geometrically consistent feature matches are strong positive evidence that two images correspond; more rarely is \emph{negative} evidence considered. 
This has remained true as deep learning has led to improvements in feature detection~\cite{detone2017toward,detone2018superpoint}, 
feature extraction~\cite{yi2016lift,ono2018lf,dusmanu2019d2,revaud2019r2d2}, and feature matching~\cite{yi2018learning,sarlin2020superglue}. 
Generally, these methods all still seek to find as many consistent correspondences as possible based on local appearance, but rarely consider global evidence that a pair of images might in reality be deceptively similar, non-overlapping views.

There are a few 
counterexamples to this view of feature matching. In bag-of-visual-words--based image retrieval methods, TF-IDF weighting is often used to downweight local features that occur across many images, because such features are poor evidence that images match~\cite{sivic2003videogoogle}. 
J\'{e}gou and Chum go further and consider \emph{missing} visual word matches as negative evidence for image-level matches~\cite{jegou2012negative}.
Other methods consider the visual change detection problem, and specifically look for differences between two views of (usually) the same 
scene~\cite{Sachdeva_WACV_2023}.
For SfM on Internet collections, a common problem 
are watermarks, timestamps, and frames (decorative borders)~\cite{weyand2015fixing,heinly2015reconstructing}.
These user-added visual elements 
yield spurious feature matches on otherwise unrelated images, which can
often be filtered with heuristics.

Most related to our work are methods for \textbf{disambiguation of image collections} in the context of SfM. These methods attempt to fix the problem of broken 3D reconstructions in the face of
repeated 
structures. 
Some SfM disambiguation methods carefully identify reliable matches~\cite{kataria2020improving},
while many others identify cues for identifying spurious matches.
A common approach is to look for evidence in the scene graph---the network of corresponding images computed during SfM---such as loops that lack geometric cycle consistency~\cite{zach2010disambiguating},
graph structures that are inconsistent with physical 3D space~\cite{wilson2013network},
and graph geodesic constraints~\cite{yan2017distinguishing}.
Other methods detect missing correspondences as a negative cue for image-level matches~\cite{zach2008can,jiang2012seeing,cui2015global}.
Heinly \etal go further and detect \emph{conflicting} 3D observations in SfM models, such as images that observe 3D points that should be impossible to see because they would be occluded by some other, spurious geometry~\cite{heinly2014correcting}. 
Other methods rely on non-visual information, like timestamps or image ordering~\cite{roberts2011structure}.
Finally, the SLAM community has studied a related problem referred to as \emph{perceptual aliasing}, where distinct locations within an environment (e.g., an office building with identical offices) yield similar visual fingerprints~\cite{angeli2008incremental,cummins2011appearance,lajoie2019modeling,ikram2022perceptual}.

Prior global methods often rely on hand-designed heuristics that can be brittle, can require manual parameter tuning for each individual reconstruction task, and can also over-segment correct 3D models. 
In contrast, our method is designed for the fundamental problem of two-frame visual disambiguation, which we solve by learning from data. 
We find that our method can disambiguate a range of SfM datasets with minimal tuning. 
It considers cues distinct from those of global image connectivity methods, such as RGB values and spatial distributions of keypoints/matches, and 
can implicitly take advantage of both positive and negative signals (like missing correspondences).
Surprisingly, we find that looking at pairs of images, without considering the full image collection, is sufficient to create correct reconstructions in nearly all of our experiments.
Furthermore, our approach is also orthogonal to global structure--based methods, 
and could naturally be combined with them.

\section{Visual disambiguation}

Our work addresses the following \emph{visual disambiguation} problem: Given two possibly very similar images, determine whether they depict the same physical 3D surface, or whether they are images of two different 3D surfaces. 
This is a binary classification task on image pairs.
A true (positive) matching pair is one where both images depict the same physical 3D surface, while a false (negative) pair observes distinct 3D surfaces with no (or very few) identical 3D points in common.
\emph{Illusory image matches}---false pairs that look similar, which we also refer to as \emph{\twinterm}---occur when two images observe distinct but visually similar 3D surfaces.

\begin{figure}
\begin{center}
\includegraphics[width=\columnwidth, trim=0 0 125 0, clip]{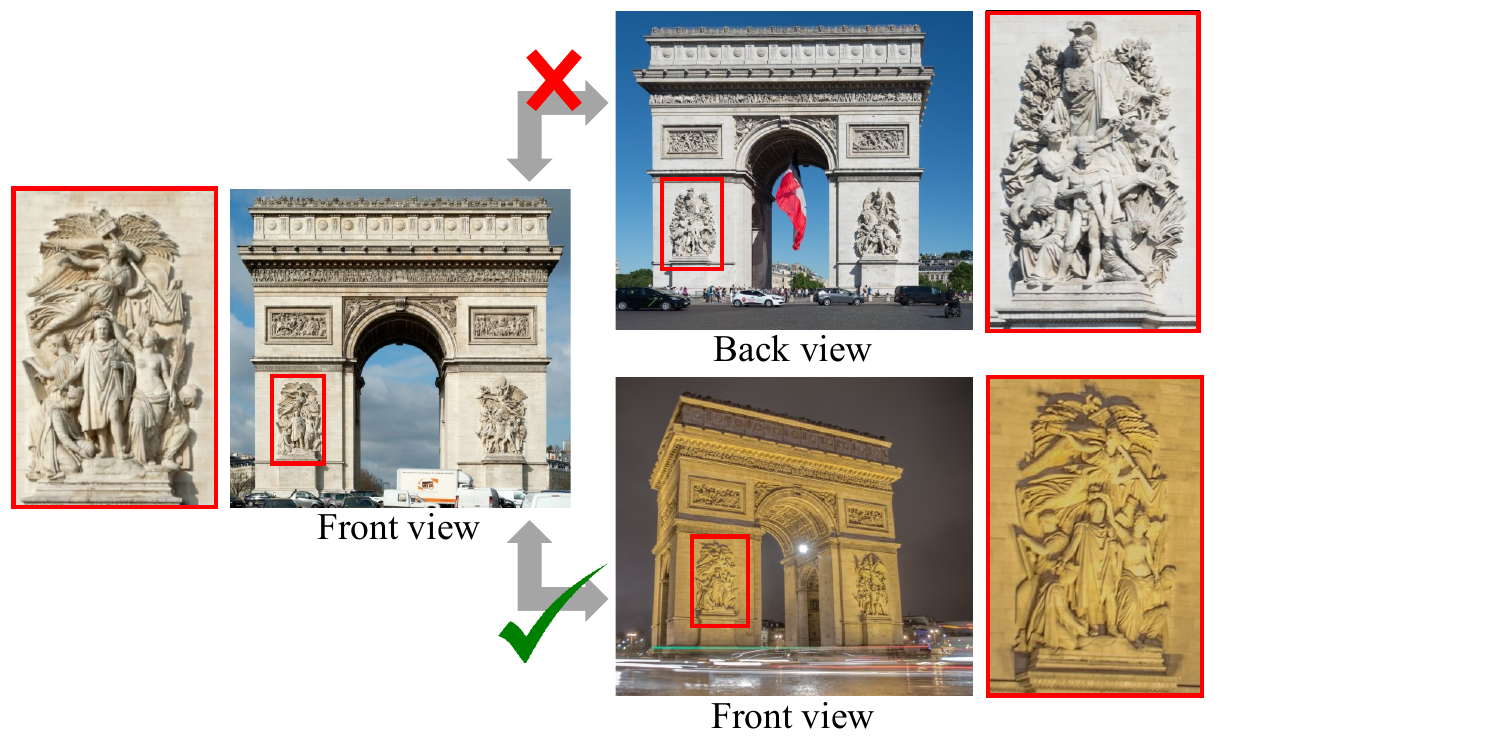}
\end{center}
\vspace{-10pt}
   \caption{
   \small
   Images captured from highly symmetric landmarks, such as the Arc de Triomphe, are difficult to disambiguate into true and false matching pairs due to repeated structures.
   However, by zooming into the sculptures, highlighted in red boxes, we can uncover subtle differences between the front and back sides, allowing us to differentiate between them.
   }
\label{fig:problem_define}
\end{figure}

Ambiguous image pairs can arise from 
symmetric buildings, repeated visual elements, and replicas of landmarks in different parts of the world.
For example, consider the images of the Arc de Triomphe shown in Figure~\ref{fig:problem_define}.
At first glance, views of the front and back of this symmetric structure appear nearly identical.
But on closer inspection we can observe differences between the two
sides, such as the distinct sculptures.
As this example illustrates, these cues can be hard to discern, as they can involve subtle differences amidst overall similar structures. 
This problem of distinguishing \twinterm is even more challenging in the presence of varying illumination, viewpoint, and transient objects.

Doppelgangers are 
also problematic in practice, especially for 3D computer vision pipelines. These often rely on feature matching methods that
may find many incorrect matches between illusory pairs. These incorrect matches 
can lead SfM methods to produce erroneous 3D reconstructions.

We find that simple classification schemes like thresholding on the number of feature matches do not suffice to identify \twinterm.
Prior image collection--level methods for visual disambiguation, discussed in Section~\ref{sec:related}, cannot disambiguate individual image pairs, and can be brittle in practice or require time-consuming parameter tuning.

In contrast, we propose a learning-based approach that trains a binary classifier on image pairs.
Since, we know of no existing dataset for visual disambiguation, we collect 
a new dataset of image pairs with carefully produced ground truth labels (Section~\ref{sec:dataset}).
Our goal is to differentiate illusory matches, even when there are only subtle differences in small regions. 
To achieve this, we propose a deep network that takes the spatial distribution of keypoint and matches as input to better reason about both local features and global and image-level information, as described in Section~\ref{sec:method}.

\section{The \ourdataset Dataset}
\label{sec:dataset}
We present the \emph{\ourdataset Dataset}, a benchmark dataset that allows for training and standardized evaluation of visual disambiguation algorithms.
\ourdataset consists of a collection of internet photos of world landmarks and cultural sites that exhibit repeated patterns and symmetric structures. 
The dataset includes a large number of image pairs, each labeled as either positive or negative based on whether they are true or false (illusory) matching pairs.
It is relatively easy to find image pairs in the wild that are correctly matching (positives). 
It is much more difficult to find and accurately label \emph{negative} image pairs.
Below, we describe how we tackle this data gathering problem.

\subsection{Mining Wikimedia Commons}
Inspired by the Google Landmarks~\cite{weyand2020google} and WikiScenes datasets~\cite{wu2021towers}, \ourdataset is collected from 
\href{https://commons.wikimedia.org/wiki/Main_Page}{Wikimedia Commons}.
Wikimedia Commons has an extensive collection of freely available images contributed by the public.
These include a large number of photos of world landmarks, organized into a hierarchy of categories.
In some landmark categories, there exists sub-categories organized according to information like viewing direction (e.g., North/South), 
providing valuable annotations for inferring geometric relationships between images.
For instance, the category consisting of exterior images of the Église de la Madeleine (a church in Paris) is called \href{https://commons.wikimedia.org/wiki/Category:Exterior_of_%C3%89glise_de_la_Madeleine}{Exterior of Église de la Madeleine}, with 
sub-categories that include \href{https://commons.wikimedia.org/wiki/Category:North_facade_of_%C3%89glise_de_la_Madeleine}{North facade of Église de la Madeleine} and \href{https://commons.wikimedia.org/wiki/Category:South_facade_of_%C3%89glise_de_la_Madeleine}{South facade of Église de la Madeleine}.
We assemble a list of landmark categories with sub-categories that contain keywords related to directions, such as ``North'', ``South'', ``East'', ``West'', ``Left'', ``Right'', ``Front'', and ``Back''. Following~\cite{wu2021towers}, we recursively download images from all sub-categories in the list to obtain an initial set of images for each landmark.
We also identified a few other visually ambiguous landmark categories for our dataset, such as replicated Disneyland castles and the Deutscher und Französischer Dom.

\subsection{Deriving image pairs with ground truth labels}
One 
approach to creating and labeling image pairs would rely solely on the directional labels on images induced by their Wikimedia Commons category. 
In particular, if two images of the same landmark are captured from the same direction (e.g., both from the north), we could label them a positive pair, whereas if they are captured from opposite directions (e.g., one from the north and one from the south), we could label them a negative pair, since images of opposite building faces are unlikely to have any true overlap.
However, this approach can produce positive pairs that lack correspondence, because two images of the same building face may not overlap (e.g., two closeups of different details).
Similar logic applies to negative pairs: we are mainly interested in \emph{hard negatives}, i.e., images of different surfaces where feature matching yields spurious correspondences.
Hence, we need a way to mine for ``interesting'' pairs.

\medskip \noindent \textbf{Finding potential doppelgangers by image matching.}
To identify interesting image pairs that share putative correspondence (correctly or incorrectly), we use the feature matching
module in COLMAP~\cite{schonberger2016structure}, a state-of-the-art SfM system.
This process yields a set of putative matching image pairs for each landmark in our dataset, along with local keypoints for each image and pairwise matches between image pairs.
We only include image pairs that have such pairwise matches in our dataset, ensuring visual similarity between the included pairs. 
For positive pairs, this means they exhibit overlap, and for negative pairs, they depict different but similar structures.

\medskip \noindent \textbf{Augmentation with image flipping.}
Some landmarks may not naturally form negative pairs because they lack similar structures when viewed from opposite directions. Therefore, it can be more difficult to find negative pairs compared to positive pairs (which are extremely abundant). 
When contemplating how to increase the number of negative training pairs, we were inspired by the image pairs like the one shown in Figure~\ref{fig:teaser}(c), where the opposite views of some buildings resemble a 2D image flip. 
Horizontally flipping one image in a pair changes it to a different, mirror-image scene~\cite{lin2020visual}, but feature matches on similar structures may still exist.
Therefore, to increase the variety of scenes in our training data, we sample a positive pair and flip one of the images, resulting in a \emph{negative} pair---a pair of similar images that nonetheless correspond to different (mirror image) surfaces.
Unlike traditional data augmentation
that generates more samples with the same label,
our approach transforms a positive pair into a negative pair.
Note that we only use such synthetic augmented pairs for training and not as test data.

\begin{figure*}[t]
\begin{center}
\includegraphics[width=\textwidth, trim=0 0 0 0, clip]{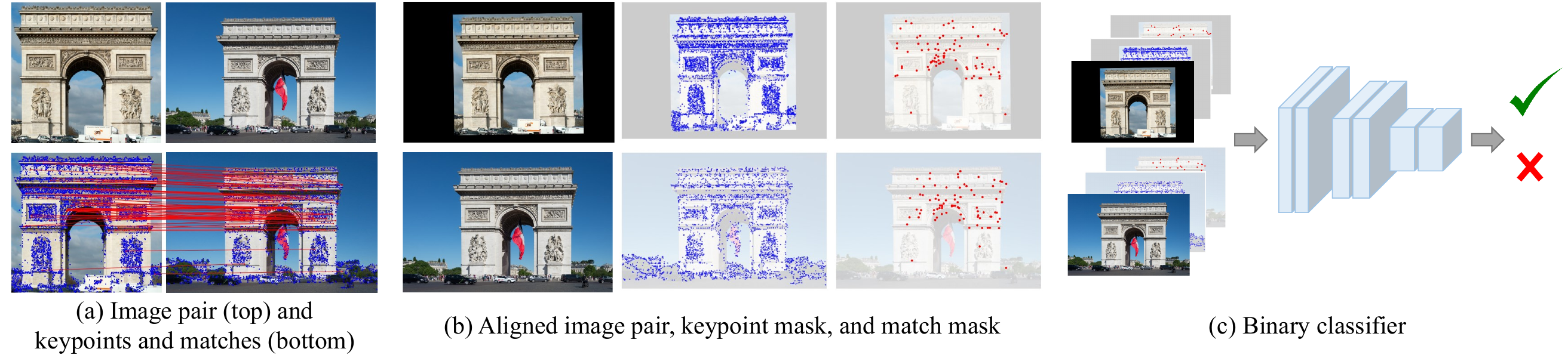}
\end{center}
\vspace{-7pt}
   \caption{
   \small
   Method overview. 
   (a) Given a pair of images, we extract keypoints and matches via feature matching methods. 
   Note that this is a negative (doppelganger) pair picturing opposite sides of the Arc de Triomphe. The feature matches are primarily in the top part of the structure, where there are repeated elements, as opposed to the sculptures on the bottom part.
   (b) We create binary masks of keypoints and matches. We then align the image pair and masks with an affine transformation estimated from matches. 
   (c) Our classifier takes the concatenation of the images and binary masks as input and outputs the probability that the given pair is positive.}    
\label{fig:method}
\end{figure*}

\subsection{Dataset statistics}
The above process results in $\sim$76K internet photos of 222 world landmarks and cultural sites, and yields over 1M visually similar image pairs. Among these pairs, 178K are labeled as negative pairs.
We provide additional data collection details and statistics in the supplemental material.

Of the 222 scenes, 58 naturally form negative pairs.
Of these 58, we split off 16 scenes (and sample 4,660 image pairs) as a test set, divided evenly into positive and negative pairs.
From the 42 scenes remaining for the training set, we sample a maximum of 3K pairs per scene to ensure balance across landmarks during training.
These 42 scenes contribute to 73K %
training pairs, again divided nearly evenly between positive and negative pairs.

Our proposed flipping augmentation on Wikimedia Commons imagery yields an additional 92K %
training pairs across 164 scenes.
To further augment negative pairs for use in training, we also applied the flipping augmentation to matching image pairs from  MegaDepth~\cite{li2018megadepth}, a large dataset of multi-view Internet photo collections. MegaDepth adds an additional 57K %
training pairs from 72 scenes.

\section{Classifying visually ambiguous pairs}
\label{sec:method}

\setlength{\abovedisplayskip}{3pt}
\setlength{\belowdisplayskip}{3pt}

We now describe how we design a classifier for visual disambiguation. 
A straightforward solution would be to train a network to take as input a raw image pair, and to output the probability that this pair is a positive match.
However, we found that this approach works poorly even starting from pre-trained models and state-of-the-art architectures~\cite{caron2021emerging, dosovitskiy2020vit}.
Visual disambiguation is a hard problem, and we conjecture that it is difficult for a network to discover all the necessary subtle cues from raw RGB images alone.

To gain insight into the problem, consider the front and back views shown in  Figure~\ref{fig:problem_define}.
The task is similar to a ``spot-the-difference'' game, where we would look for image regions that \emph{should} correspond, but don't.
For instance, in Figure~\ref{fig:problem_define}, the sculptures should match across the 
views, but are different.
We might also note mismatched structures in the background or even on low-level details like individual bricks.
To allow the network to perform similar reasoning, we provide it with information about the spatial distribution of distinctive keypoints and keypoint matches.
In addition, we perform a rough alignment of the images to allow the network to directly compare corresponding regions.
We describe these enhancements below.

\subsection{Spatial distribution of keypoints and matches}
\label{subsec:match}
Given two input images, as a pre-processing step, we compute keypoint matches between them, then use RANSAC~\cite{fischler1981random} to estimate a Fundamental matrix and filter out outlier matches.
We provide the locations of all detected keypoints, as well as all (filtered) matches as an additional network input in the form of two binary mask images.

The idea is that the keypoint and match locations provide useful signals to the network. For instance, by showing the network where keypoint matches \emph{were} found, it also lets the network know where matches \emph{weren't} found (where there are keypoints but no matches). Such regions may indicate missing or distinct objects. The network can also compute other signals if it chooses, like the raw number of matches (a reasonable baseline for visual disambiguation).
As an example, we illustrate SIFT keypoints and matches 
for a doppelganger pair in Figure~\ref{fig:method}(a). We see that matches are denser in regions with visually similar structures, but much sparser in regions with distinct structures such as sculptures.

In particular, for an image pair $(I_a,I_b)$ (of dimensions $H \times W$), we extract keypoints and matches.
We denote the set of keypoints for image $I_a$ as 
$\mathcal{K}_a=\{\mathbf{x}_i^a\}_{i=1}^{N_a}$, 
where $\mathbf{x}_i^a$ is the pixel location of the $i^\text{th}$ keypoint.
Similarly, we denote matches between this image pair as a set of keypoint pairs
$\mathcal{M}_{a,b}=\{(\mathbf{x}^a_{i_k}, \mathbf{x}^b_{j_k})\}_{k=1}^{M_{a,b}}$.
We create a binary mask of keypoints $\mathbf{K}_a\in \{0,1\}^{H\times W}$ 
using the keypoints $\mathcal{K}_a$, 
where pixels corresponding to keypoints (rounded to grid location) are set to one, and other pixels are set to zero.
Similarly, we create a binary mask of matches $\mathbf{M}_{a,b}^a \in \{0,1\}^{H\times W}$ for $I_a$, 
where pixels corresponding to matches are set to one, and all other pixels are set to zero.
These keypoint and match masks are illustrated in Figure~\ref{fig:method}(b).
Our classifier 
takes these masks as input, along with the RGB image pair.

\medskip \noindent \textbf{Alignment for better comparison.}
\label{subsec:align}
To facilitate comparison of potentially corresponding regions, we also perform a rough geometric alignment of the input image pair. 
We estimate an affine transform $T$
from the matches $\mathcal{M}_{a,b}$,
and warp the images and the binary masks accordingly. Figure~\ref{fig:method}(b) shows an example. Note that the alignment need not be perfect; the goal is simply to bring regions that the network might wish to compare closer together.

\subsection{Binary classification}
As illustrated in Figure~\ref{fig:method}(c), 
our classifier $F_\theta$ takes an image pair and derived binary keypoint and match masks 
as input, and outputs the probability that the pair is a positive match. 
We concatenate the RGB images and masks for both images into a $6+4=10$-channel input tensor.
To optimize our classifier $F_\theta$, we use a focal loss~\cite{lin2017focal}, a modified version of the cross-entropy loss. The focal loss improves performance by balancing the distribution of positive and negative pairs, and giving more weight to hard examples.

\subsection{Implementation details}

\noindent \textbf{Keypoint and match masks.}
After resizing and padding each image to $1024\times1024$ resolution, 
we compute matches using LoFTR~\cite{sun2021loftr}, a detector-free, learned local feature matching method.
We use a LoFTR model pretrained on MegaDepth~\cite{li2018megadepth}, which focuses on outdoor scenes.  
We then perform geometric verification using RANSAC~\cite{fischler1981random}, and establish the keypoint mask using all LoFTR output matches and the match mask using geometrically verified matches.

\medskip \noindent \textbf{Network and training.}
The classifier $F_\theta$ consists of 3 residual blocks~\cite{he2016deep}, an average pooling layer, and a fully connected layer.
We train for 10 epochs with a batch size of 16 using the Adam optimizer with an initial learning rate of $5\times10^{-4}$. The learning rate is linearly decayed to $5\times10^{-6}$ from epoch 5 onwards.
Additional implementation details are provided in the supplemental material.

\begin{table*}[t]
\small
\centering
\resizebox{0.95\textwidth}{!}{

\begin{tabular}{lccccccc}
\toprule
                                                                                      & \multicolumn{2}{c}{SIFT~\cite{lowe2004distinctive}+RANSAC~\cite{fischler1981random}} & \multicolumn{2}{c}{LoFTR~\cite{sun2021loftr}} & \multicolumn{2}{c}{DINO~\cite{caron2021emerging}-ViT} & \multirow{2}{*}{Ours} \\
                                                                                      & \#matches                                             & \multicolumn{1}{c}{\%matches}                                            & \#matches              & \multicolumn{1}{c}{\%matches}              & \multicolumn{1}{c}{Latent code}      & \multicolumn{1}{c}{Feature map}      &                       \\ \midrule
Average precision     & 83.4 & 81.2 & 85.3 & 86.0 & 62.0 & 63.3 & \textbf{95.2} \\ 
ROC AUC     & 80.2 & 77.1  & 78.9 & 80.3 & 60.9 & 61.5  & \textbf{93.8}      
\\ \midrule
Alexander Nevsky Cathedral, Łódź           & 72.7 & 75.9 & 80.7 & 80.4 & 50.9 & 50.3 & \textbf{89.5} \\
Alexander Nevsky Cathedral, Sofia          & 89.5 & 87.6 & 90.0 & 92.2 & 53.0 & 53.6 & \textbf{98.5} \\
Alexander Nevsky Cathedral, Tallinn        & 73.1 & 76.0 & 76.1 & 80.3 & 58.8 & 50.8 & \textbf{86.2} \\
Arc de Triomphe                            & 86.1 & 81.7 & 85.7 & 93.3 & 55.4 & 61.1 & \textbf{97.6} \\
Berlin Cathedral                           & 91.8 & 91.6 & 93.6 & 92.7 & 76.4 & 70.6 & \textbf{99.4} \\
Brandenburg Gate                           & 79.3 & 73.7 & 90.9 & 95.6 & 60.8 & 60.9 & \textbf{99.8} \\
Cathedral of Saints Peter and Paul in Brno & 95.8 & 96.4 & 89.8 & 88.4 & 64.6 & 79.9 & \textbf{99.8} \\
Cathedral of St Alexander Nevsky, Prešov   & 82.5 & 74.0 & 86.1 & 85.3 & 62.9 & 64.8 & \textbf{94.6} \\
Charlottenburg Palace                      & 81.5 & 76.1 & 85.6 & 81.1 & 65.8 & 54.1 & \textbf{93.3} \\
Church of Savior on the Spilled Blood      & 82.1 & 73.2 & 84.9 & 75.5 & 63.9 & 67.5 & \textbf{93.8} \\
Deutscher und Französischer Dom (Berlin)   & 74.5 & 71.9 & 85.8 & 84.2 & 55.6 & 51.5 & \textbf{98.1} \\
Florence Cathedral                         & 90.6 & 83.8 & 84.5 & 82.0 & 54.6 & 63.8 & \textbf{94.2} \\
Sleeping Beauty Castle                     & 81.1 & 81.2 & 75.0 & 85.6 & 67.2 & 66.4 & \textbf{97.1} \\
St.\ Vitus Cathedral                        & 96.8 & 88.0 & 89.2 & 87.5 & 84.0 & 77.0 & \textbf{99.8} \\
Sydney Harbour Bridge                      & 79.4 & \textbf{92.3} & 83.8 & 86.2 & 53.0 & 75.5 & 87.0 \\
Washington Square Arch                     & 77.7 & 75.9 & 82.8 & 86.0 & 65.2 & 65.0 & \textbf{95.1}
\\ \bottomrule
\end{tabular}
}
\vspace{7pt}
\caption{
Quantitative results for visual disambiguation evaluated on the \ourdataset test set. 
The first two rows show the average precision and ROC AUC multiplied by 100, respectively, averaged over the 16 test scenes.
The remaining rows show the average precision, multiplied by 100, for each individual scene.
\emph{\#matches} refers to thresholding the number of matches and \emph{\%matches} refers to thresholding the ratio of the number of matches to the number of keypoints, as described in Section \ref{sec:visual_result}.
}
\label{tab:quantitative}
\end{table*}
\section{Experiments}
\label{sec:experiment}

In this section, we evaluate our visual disambiguation method on the \ourdataset dataset. 
We then discuss how our pairwise classifier can be integrated into a SfM pipeline for reconstructing visually ambiguous image collections.
Our experimental results demonstrate the effectiveness and generalization power of our method for SfM disambiguation. 
Finally, we provide an ablation study to validate the 
effectiveness of each component in our network design.

\subsection{Visual disambiguation}
\label{sec:visual_result}

\noindent \textbf{Baselines.}
We compare our method to three baselines:

One set of baselines %
simply uses the number of local feature matches to predict if an image pair is a positive (true) match. 
We 
evaluate two feature matching baselines, one based on SIFT~\cite{lowe2004distinctive} and one on LoFTR~\cite{sun2021loftr}.
Both SIFT and LoFTR matches are filtered via geometric verification using RANSAC, yielding a cleaner set of matches for use in classification.
Our baselines use these matches in two ways: (1) thresholding the number of matches after geometric verification, and (2) thresholding the ratio of the number of matches to the number of keypoints. The idea behind (2) is that if there are few matches relative to the number of keypoints, that may be a sign that the pair is a doppelganger.

In addition, we compare our method to DINO~\cite{caron2021emerging}, 
self-supervised features that achieve state-of-the-art image classification and semantic segmentation results.
We train a classifier using on the latent codes and feature maps produced by the pretrained DINO model.

\medskip  \noindent \textbf{Quantitative results.}
Table~\ref{tab:quantitative} presents quantitative results for visual disambiguation evaluated on the \ourdataset test set,
reported in terms of average precision (AP) and ROC AUC score, averaged across the 16 test scenes.
Our method outperforms all other baselines with an AP of 95.2\% and ROC AUC of 93.8\%.
DINO produces much poorer results, possibly because it generates features that are well-suited for semantic classification tasks but not for visual disambiguation.
For feature matching methods, 
we find that the number (or ratio) of matches alone is insufficient to perform well on this task.
Our method can leverage not only the number of matches, but also rich information about the spatial distribution of keypoints and matches.

\begin{figure}
    \centering
    
    \includegraphics[width=\columnwidth, trim=0 30 0 0, clip]{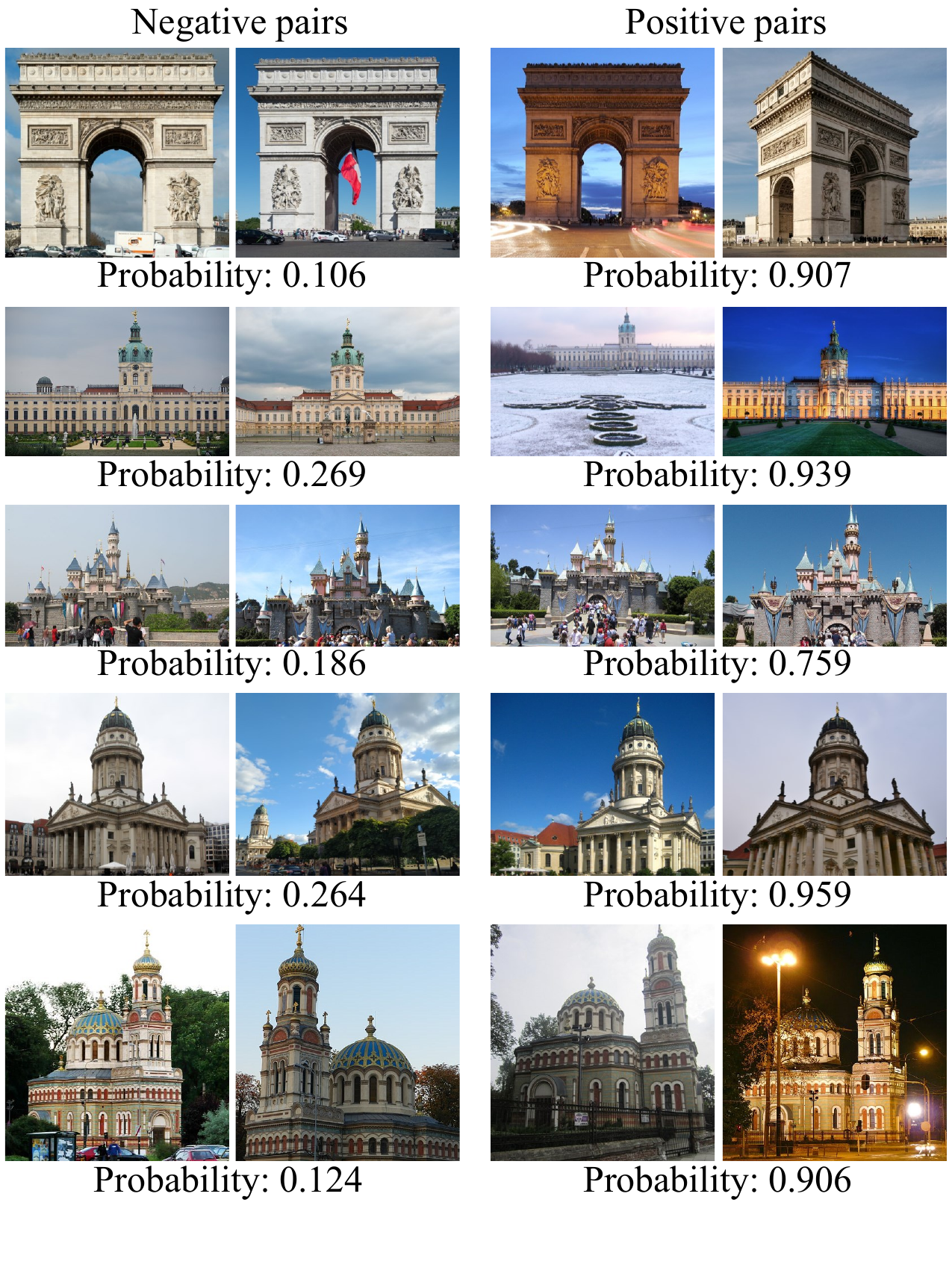}
    \vspace{-10pt}
    
   \caption{ 
   \small   Visual disambiguation results. 
   We display test pairs and their corresponding probability of being a positive match, as predicted by our network.
   Negative pairs are shown in the left column and positives in the right column. Note that our network cleanly separates the negative and positive pairs by score, including in the presence of varying illumination and other factors.
   }
\label{fig:result}
\end{figure}

\medskip  \noindent \textbf{Visualizing test pairs.}
We show a visualization of test pairs along with our network's prediction scores in Figure~\ref{fig:result}. 
The test set covers a variety of scenes and includes different types of visual ambiguity, such as symmetric structures, replicas of landmarks, and twin buildings. 
The pairs on the left of the figure are doppelgangers that are visually challenging to distinguish. 
Our network confidently predicts them as negative pairs.
On the right of the figure, we show positive pairs with varying viewpoint and illumination. Our method correctly recognizes them as depicting the same 3D surfaces.

Further details on the baseline implementations,
additional comparisons with D2-Net+RANSAC~\cite{dusmanu2019d2,fischler1981random} and
SuperPoint+SuperGlue~\cite{detone2018superpoint,sarlin2020superglue}, 
more quantitative results (including per-scene results, confusion matrices, precision-recall curves, and false positive rates),
and failure cases discussion 
are provided in the supplemental material.

\subsection{Structure from motion disambiguation}
We integrate our binary classifier into COLMAP's SfM pipeline~\cite{schonberger2016structure} to evaluate its use in disambiguating scenes with duplicate and symmetric structures.

\medskip  \noindent \textbf{Benchmark data.}
We evaluate our method's use on the SfM problem on benchmark datasets from Heinly et al.~\cite{heinly2014correcting}, Wilson et al.~\cite{wilson2013network}, and Roberts et al.~\cite{roberts2011structure}, along with several datasets we collected from Flickr.
Altogether, we consider two kinds of datasets: 
Internet photo collections of landmarks (13 
landmarks that are difficult to reconstruct due to symmetric and repeated structures), and 
three non-landmark collections with repeated or duplicate structures from \cite{roberts2011structure}
that significantly differ from our training data.
Our model is primarily trained on landmark collections, but we include the non-landmark scenes from~\cite{roberts2011structure} to evaluate its generalization ability to different types of collections.
Table~\ref{tab:sfm_comparison} shows the number of images in each scene.
These scenes are not present in our training data.

\medskip  \noindent \textbf{Integrating our method into SfM.}
SfM takes a collection of images $\mathcal{I}=\{I_i\}_{i=1}^n$ 
as input.
A feature extraction and matching stage first produces a set of image pairs with 
geometrically verified matches, with these pairs denoted as $\mathcal{P}=\{(I_a,I_b)|a,b\in \{1,\dots,N_{\mathcal{I}}\}\}$. 
A scene graph $\mathcal{G}=(\mathcal{I}, \mathcal{P})$ is then established, with images as nodes and image pairs as edges.
A reconstruction stage then takes $\mathcal{G}$ as input and computes camera poses for a subset of
images, along with a 3D point cloud.

Illusory image pairs with repeated and symmetric structures can produce spurious matches, leading to broken reconstructions.
To detect and remove the illusory pairs, we use our binary classifier as a filter on the edges of the scene graph.
Our classifier outputs a probability that an image pair $(I_a,I_b)$ in $\mathcal{P}$ is a positive pair, which we can then threshold to remove edges from the scene graph with probability lower than a threshold $\tau$.
If the classifier removes all incorrect edges (and only those edges), 
SfM can produce a disambiguated, correct reconstruction.
However, if the classifier removes too few edges, the reconstruction may still be incorrect; if it removes too many edges, the scene graph may break apart and result in several partial reconstructions.
Hence, this task is a good test for the performance of a pairwise classifier.

\medskip  \noindent \textbf{Baselines.}
We compare our method with several baselines, including ``vanilla'' COLMAP~\cite{schonberger2016structure}, which is a state-of-the-art SfM system.
The default threshold for the number of matches for a valid image pair in COLMAP is 15. 
We also consider the simple baseline of setting a much higher threshold of 150 matches to remove potential doppelganger pairs.
Additionally, we compare our method to four SfM disambiguation methods:
Heinly et al.~\cite{heinly2014correcting} propose a post-processing method that analyzes reprojected geometry conflicts between images.
Wilson et al.~\cite{wilson2013network} prunes bad tracks, while Cui et al.~\cite{cui2015global} and Yan et al.~\cite{yan2017distinguishing} filter out incorrect matches between images prior to SfM. 
These methods are based on heuristics that consider a collection of images at a time.
We utilize the default hyperparameters provided in the GitHub implementations of \cite{heinly2014correcting}\footnotemark\footnotetext{\url{https://github.com/jheinly/sfm_duplicate_structure_correction}} and \cite{wilson2013network}\footnotemark\footnotetext{\url{https://github.com/wilsonkl/sfm-disambig}}, respectively.
For \cite{cui2015global,yan2017distinguishing}, we follow the implementation 
provided in the COLMAP disambiguation GitHub repository\footnotemark\footnotetext{\url{https://github.com/cvg/sfm-disambiguation-colmap}},
and use hyperparameters tuned for the Alexander Nevsky Cathedral landmark on other landmark scenes, and hyperparameters tuned for the Street scene on other datasets from \cite{roberts2011structure}.
For our method, we use the same probability threshold across all landmarks, without tuning per scene.

\begin{table*}
\centering
\small
\resizebox{\textwidth}{!}{
\begin{tabular}{llcccccccc}
\toprule
                                          &                                                         & Images & COLMAP ~\cite{schonberger2016structure} & ~\cite{schonberger2016structure} \#matches$>$150 & Heinly et al.~\cite{heinly2014correcting} & Wilson et al.~\cite{wilson2013network} & Cui et al.~\cite{cui2015global} & Yan et al.~\cite{yan2017distinguishing} & Ours       \\
                                          \midrule
\parbox[t]{2mm}{\multirow{13}{*}{\rotatebox[origin=c]{90}{Internet landmarks}}}
& Alexander Nevsky Cathedral~\cite{heinly2014correcting}  & 448    & \ding{55}                               & \ding{55}                                        & \ding{51}                                 & \ding{55}                              & \ding{51}                       & \ding{51}                               & \ding{51}  \\
                                          & Arc de Triomphe~\cite{heinly2014correcting}             & 434    & \ding{55}                               & \ding{55}                                        & \ding{51}                                 & \ding{55}                              & \ding{55}                       & \ding{51}                               & \ding{51}  \\
                                          & Berliner Dom~\cite{heinly2014correcting}                & 1,618  & \ding{55}                               & \ding{51}                                        & \ding{51}                                 & \ding{55}*                             & \ding{51}                       & \ding{55}*                              & \ding{51}  \\
                                          & Big Ben~\cite{heinly2014correcting}                     & 402    & \ding{55}                               & \ding{55}                                        & \ding{51}                                 & \ding{55}                              & \ding{51}                       & \ding{55}                               & \ding{51}  \\
                                          & Brandenburg Gate~\cite{heinly2014correcting}            & 175    & \ding{55}                               & \ding{51}                                        & \ding{51}                                 & --                                     & \ding{55}                       & \ding{55}                               & \ding{51}  \\
                                          & Church on Spilled Blood~\cite{heinly2014correcting} & 277    & \ding{55}                               & \ding{55}                                        & \ding{51}                                 & --                                     & \ding{55}                       & \ding{55}                               & \ding{55}  \\
                                          & Radcliffe camera~\cite{heinly2014correcting}            & 282    & \ding{55}                               & \ding{51}                                        & \ding{51}                                 & \ding{55}*                             & \ding{51}                       & \ding{51}                               & \ding{51}  \\
                                          & Sacre Coeur~\cite{wilson2013network}                    & 5,450  & \ding{55}                               & \ding{55}                                        & \ding{51}$^\dagger$                                 & \ding{55}*                              & --                              & \ding{55}*                              & \ding{55}  \\
                                          & Seville~\cite{wilson2013network}                        & 2,396  & \ding{55}                               & \ding{51}                                        & \ding{55}                                 & \ding{51}                              & --                              & \ding{55}*                              & \ding{51}  \\
                                          & Florence Cathedral [Flickr]                             & 8,674  & \ding{55}                               & \ding{55}                                        & --                                        & \ding{55}                              & --                              & \ding{51}                               & \ding{51}  \\
                                          & St. Vitus Cathedral [Flickr]                            & 5,059  & \ding{55}                               & \ding{55}                                        & \ding{51}                                 & \ding{51}                              & --                              & \ding{55}*                              & \ding{51}  \\
                                          & Temple of Heaven [Flickr]                               & 1,538  & \ding{55}                               & \ding{55}                                        & \ding{55}                                 & \ding{55}                              & --                              & \ding{55}*                              & \ding{51}  \\
                                          & York Minster [Flickr]                                   & 3,902  & \ding{55}                               & \ding{51}                                        & \ding{55}                                 & \ding{55}                              & --                              & \ding{55}                               & \ding{51}  \\
                                          \midrule
\parbox[t]{2mm}{\multirow{3}{*}{\rotatebox[origin=c]{90}{Others}}}
& Cereal~\cite{roberts2011structure}                      & 25     & \ding{55}                               & \ding{55}                                        & \ding{51}                                 & \ding{55}                              & \ding{55}                       & \ding{51}                               & \ding{55}  \\
                                          & Cup~\cite{roberts2011structure}                         & 64     & \ding{55}                               & \ding{55}                                        & \ding{55}                                 & \ding{55}                              & \ding{55}                       & \ding{51}                               & \ding{51}  \\
                                          & Street~\cite{roberts2011structure}                      & 19     & \ding{55}                               & \ding{55}                                        & \ding{55}                                 & \ding{55}                              & \ding{51}                       & \ding{51}                               & \ding{55}* \\ 
                                          \midrule
\multicolumn{2}{l}{Number of scenes: \ding{51}/\ding{55}*/\ding{55}}                                &        & 0/0/16                                  & 5/0/11                                           & 10/0/5                                    & 2/3/9                                  & 5/0/5                           & 7/5/4                                   & 12/1/3    
\\
\bottomrule
\end{tabular}
}
\vspace{7pt}
\caption{
Structure from Motion disambiguation results. 
\ding{51} means that a scene is correctly disambiguated and reconstructed. \ding{55} means that a method fails to disambiguate the scene, and \ding{55}* means a scene is over-split.
`--' means that a method fails to produce a reconstruction. 
\cite{heinly2014correcting} fails to generate a reconstruction for Florence Cathedral ($>$8k images) due to memory issues.
\cite{wilson2013network} requires focal length information, which is unavailable for the Brandenburg Gate and Church on Spilled Blood datasets.
\cite{cui2015global} fails to produce results on large-scale scenes due to numerical errors.
With a single threshold, our method successfully reconstructs 12 out of 16 scenes.
$^\dagger$The Sacre Coeur result from~\cite{heinly2014correcting} uses a different subset of images as reported in their paper.
}
\label{tab:sfm_comparison}
\end{table*}

\medskip  \noindent \textbf{Reconstruction results.}
The reconstruction results are summarized in Table~\ref{tab:sfm_comparison}.
SfM successes and failures are identified by checking for conflicts between the reconstruction and the corresponding images or 3D mesh from Google Earth.

Our method successfully disambiguates and reconstructs 12 out of 16 scenes all using the same parameter settings, achieving the highest number of correctly reconstructed scenes of all methods.
We also evaluated our method using several probability thresholds, and found the results to be robust to the setting of the threshold (more results in supplemental).
The generalization ability of our learning-based method is evident from the fact that it can be applied to new scenes that haven't been observed during training, without the need for fine-tuning or parameter tuning.

Our method fails on a few test scenes with the default threshold, but we found our method can successfully disambiguate through threshold tuning. 
No method, except \cite{heinly2014correcting}, reconstructs the Church on Spilled Blood correctly.
However, we found that with a higher score threshold, our method correctly splits this scene into sub-models (as there is insufficient overlap in the input images to produce a single unified model).
This suggests that the scene is particularly challenging to disambiguate, requiring a more stringent threshold to filter out illusory image pairs. 
We observe a similar pattern with the Sacre Coeur dataset.
Our method with default settings fails on the Cereal and Street datasets from~\cite{roberts2011structure}, which are quite different from our training scenes. 
However, we found that our method can accurately reconstruct without over-splitting on both scenes by tuning the threshold.
These results imply that probabilities predicted by our model maintain a reasonable ordering, with lower probabilities assigned to illusory image pairs and higher probabilities to positive pairs. 
The need for tuning is likely due to the difference in domain between our training data and the datasets from~\cite{roberts2011structure}.

\begin{figure*}
\begin{center}
\includegraphics[width=0.99\textwidth, trim=5 10 5 0, clip]{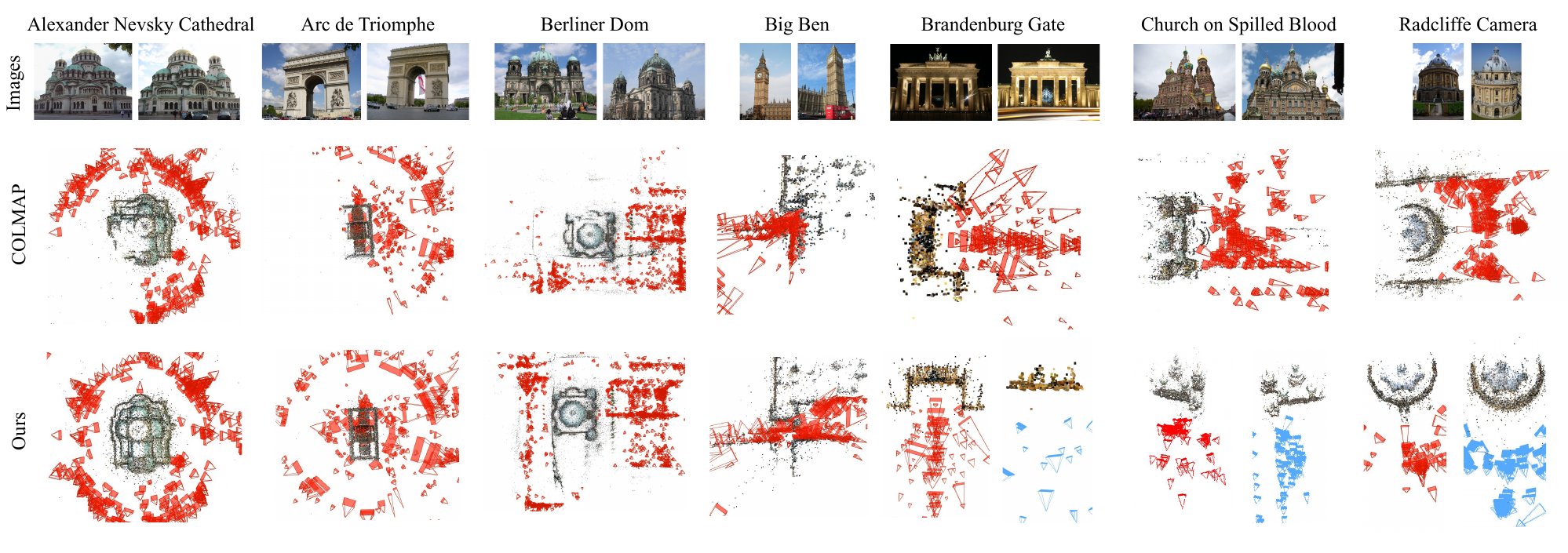}
\includegraphics[width=0.99\textwidth, trim=0 0 0 0, clip]{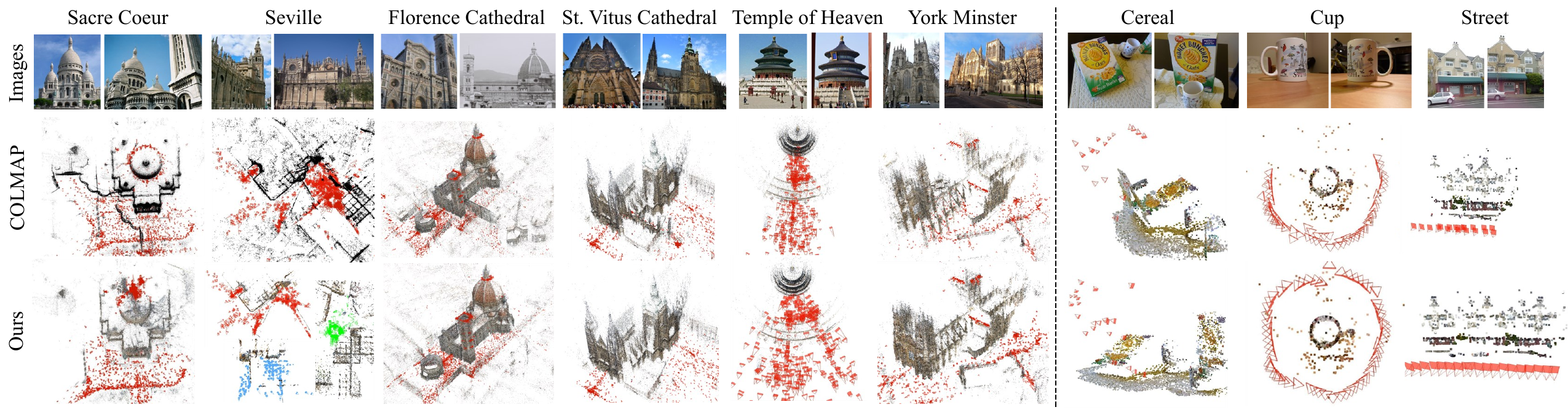}
\end{center}
\vspace{-5pt}
   \caption{
   \small
   Visualization of Structure from Motion disambiguation results. 
   We show a doppelganger pair at the top of each scene,
   vanilla COLMAP reconstructions in the middle,
   and our method's disambiguated reconstructions at the bottom.
   Note that for some landmarks, the correct reconstruction is separated into multiple components when disambiguated due to a lack of camera views from sufficient viewpoints (shown as multiple submodels with red, blue, and green cameras).
   Our results are generated using the same threshold on the image match probabilities, except for the Church on Spilled Blood, Cereal, and Street datasets.
   }
\label{fig:colmap7scenes}
\end{figure*}

We show reconstructions produced by vanilla COLMAP and those disambiguated with our method in Figure~\ref{fig:colmap7scenes}.
Vanilla COLMAP yields ghost structures such as extra towers, domes, and facades for landmarks like Alexander Nevsky Cathedral, Berliner Dom, Church on Spilled Blood, Sacre Coeur, Seville, Florence Cathedral, and St.\ Vitus Cathedral. 
It produces reconstructions that collapse to one side for landmarks like the two-way symmetric Arc de Triomphe and Brandenburg Gate, the four-way symmetric Big Ben and the tower of York Minster, the dome of Radcliffe Camera, and the highly symmetric Temple of Heaven.
Our method can disambiguate the range of ambiguities that appear in these scenes, resulting in correct COLMAP reconstructions.

\subsection{Ablation study}

\begin{table}
\centering
\small
\begin{tabular}{llc}
\toprule
\multirow{2}{*}{Backbone} &  CNN (Full)        & 95.2 \\
                          &    Transformer       & 95.5 \\ \midrule
Dataset                       & w/o Augmentation          & 93.6 \\\midrule
\multirow{3}{*}{Design}        & w/o Masks       & 64.7 \\
                              & SIFT+RANSAC masks & 87.6 \\
                              & w/o Alignment       & 92.3 \\\bottomrule
\end{tabular}

\vspace{7pt}
\caption{Ablation study of backbone selection, dataset augmentation, and network input design. Results are reported as average precision times 100.
}
\label{tab:ablation}
\end{table}
We conduct an ablation study to evaluate 
three factors: backbone selection, dataset augmentation, and network input design. 
The results, reported as average precision scores, are shown in Table~\ref{tab:ablation}.
Additional ablations on combination of factors are provided in the supplemental material.

In the \emph{Transformer} experiment, we replace the CNN backbone with a vision transformer; we find that this setting achieves comparable results, but with longer training time.
In the \emph{w/o Augmentation} experiment, we train the classifier on 42 scenes without flip augmentations. This setting results in lower performance, indicating that our flip augmentation effectively increases the amount of useful training data.

The remaining variations are trained on the dataset of 42 scenes without augmentation for speed of training.

\medskip  \noindent \textbf{Keypoint and match masks.}
In the \emph{w/o Masks} setting, the keypoint and match masks are removed from the input, leaving only RGB images.
This significantly degrades visual disambiguation performance, with a drop in average precision from 93.6\% to 64.7\%, validating the importance of the keypoint and match masks as network inputs.
In the \emph{SIFT+RANSAC masks} experiment, the LoFTR keypoint and match masks are replaced with masks computed with SIFT+RANSAC, leading to degraded performance due to the lower quality of SIFT+RANSAC masks compared to those from LoFTR. 
However, this version still outperforms other baselines.

\medskip  \noindent \textbf{Alignment.}
In the \emph{w/o Alignment} experiment, we do not align the input pair, resulting in a decrease in average precision from 93.6\% to 92.3\%.

\section{Conclusion}
We tackle the visual disambiguation problem by framing it as a binary classification task on image pairs.
We propose a learning-based approach,
and collect a new dataset, \ourdataset, which consists of visually similar image pairs with binary labels.
We design a classification network that leverages local features and matches.
Our experiments show that our method outperforms baselines and alternative network designs, 
achieving surprisingly good performance on this challenging disambiguation task.
Furthermore, we integrate our learned classifier into an SfM pipeline and show that it can produce correctly disambiguated reconstructions on difficult landmarks.

\smallskip  \noindent \textbf{Acknowledgements.}
We thank Zhengqi Li, David Fouhey, and Bill Freeman for their valuable discussions. 
This work was supported in part by a Snap Research Fellowship and by the National Science Foundation (IIS-2008313).

{\small
\bibliographystyle{ieee_fullname}
\bibliography{ref}
}

\clearpage
\appendix
\section{The \ourdatasetfull}

\subsection{Data collection process}
The process of creating image pairs with ground truth labels posed several challenges, including the difficulty of finding potential doppelgangers (as described in the main paper) and dealing with erroneously categorized images on Wikimedia Commons. 
Such images can lead to incorrect labels for image pairs that include them, which can affect the quality of our dataset.
To address this issue, we propose to use a K-NN (K-Nearest Neighbor) algorithm to identify those images and ensure that \ourdatasetfull comprises high-quality image pairs of similar structures with accurate labels.

\smallskip \noindent \textbf{Identifying incorrectly categorized images.}
While we find the most Wikimedia Commons images are correctly categorized, we found that some images are uploaded to the wrong subcategory, perhaps because people can themselves be confused about what side of a symmetric building they are looking at.
Unfortunately, even a single incorrectly labeled image can lead to a large number of incorrect negative pairs that have many feature matches (because in reality they should be positive pairs). 
Therefore, to avoid noisy labels in our dataset,
we must identify and remove such incorrectly categorized images.
For this, we look at the scene graph computed by COLMAP~\cite{schonberger2016structure} and remove images whose label is different from other images with a similar connectivity pattern in the scene graph.

Specifically, we use the K-NN (K Nearest Neighbor) algorithm~\cite{cover1967nearest} to identify such images, based on the similarity of connectivity computed from the scene graph. 
First, we construct an adjacency matrix $A$ where each element $A(i,j)$ represents the number of matches between image $i$ and image $j$.
Next, we normalize the connectivity vector for each image to a unit vector, where the connectivity vector of image $i$  is the $i^{th}$ row vector of adjacency matrix $A$.
We calculate the similarity of connectivity between any two images as the dot product of their respective connectivity vectors.
Suspicious images are identified as those with different labels from their neighbors, and we remove pairs containing such images from our dataset.

\subsection{Dataset Statistics}

Table~\ref{tab:train_set} and Table~\ref{tab:test_set} provide additional statistics on the \ourdataset training and test sets. The tables list the test scenes and training scenes that naturally form negative pairs, along with the average and the 95th percentile number of matches per scene. 
Our dataset includes a variety of landmarks, such as cathedrals, museums, castles, and other notable structures. The exteriors of these landmarks exhibit repeated and symmetric patterns. Most scenes in both the training and test sets average more than 50 matches.

\section{Visual disambiguation}

\subsection{Implementation details of our method}

\medskip \noindent \textbf{Keypoint and match masks.}
Given a pair of images, we resize and pad them to a resolution of $1024\times1024$.
We then use LoFTR~\cite{sun2021loftr}, a learning-based feature matching method, to match the image pair.
LoFTR produces matches and scores for each match.
We filter out weak matches by applying a threshold of 0.8 to the scores.
To further refine matches, we perform geometric verification by estimating the fundamental matrix using RANSAC~\cite{fischler1981random} with a reprojection error of 3 and a confidence level of 0.99.
For this step, we use the publicly available OpenCV implementation.
We use all the output matches to establish keypoint masks, and the geometrically verified matches to establish match masks.

\medskip \noindent \textbf{Input alignment.}
After obtaining the keypoints and matches, we estimate an affine transformation matrix using the OpenCV implementation of RANSAC with a inlier error of 20 pixels.
We set a larger threshold, which means that an affine transform will only roughly fit the data, because we need a more tolerant threshold to have enough inliers to fit a transform at all.
We use the estimated affine transformation matrix to align the images, keypoint masks, and match masks.

\medskip \noindent \textbf{Network architecture.}
Our network architecture and parameter settings are similar to ResNet-18~\cite{he2016deep}, but we use three residual blocks with channel dimensions of 128, 256, and 512. After the average pooling layer, the last fully connected layer takes a 512-dimensional input and outputs a 2-dimensional vector. We then apply softmax to the vector to obtain probabilities.

\medskip \noindent \textbf{Training.}
We train our network for 10 epochs using a batch size of 8 with two NVIDIA GeForce RTX 2080 Ti GPUs. 
The training process took approximately 9 hours for the 42 scenes and 30 hours for all scenes with image flipping augmentation.
For optimization, we used the Adam optimizer with parameters $\beta_1 =0.9$ and $\beta_2 = 0.999$, an initial learning rate of $5\times10^{-4}$, and linearly decayed the learning rate starting at epoch 5 until it reaches $5\times10^{-6}$ at epoch 10.

\begin{table}[]
\begin{tabular}{lcc}

\toprule
Training scene                & Mean & 95\%   \\\midrule
Aleppo Citadel                 & 90   & 313  \\
Almudena Cathedral             & 81   & 298  \\
Arc de Triomphe du Carrousel   & 88   & 317  \\
Brooklyn Bridge                & 47   & 152  \\
Château de Chambord            & 96   & 344  \\
Château de Cheverny            & 75   & 284  \\
Château de Sceaux              & 134  & 629  \\
Cinderella Castle              & 148  & 721  \\
Cour Carrée (Louvre)           & 129  & 475  \\
Cour Napoléon                  & 94   & 295  \\
Da Lat Station                 & 84   & 253  \\
Église de la Madeleine         & 119  & 390  \\
Eiffel Tower                   & 62   & 197  \\
El Escorial                    & 132  & 452  \\
Grande Galerie (Louvre)        & 99   & 324  \\
Grands Guichets du Louvre      & 140  & 321  \\
Liberty Square, Taipei         & 65   & 197  \\
London Eye                     & 56   & 147  \\
Mainz Cathedral                & 116  & 234  \\
Market Square in Wrocław       & 118  & 429  \\
Notre-Dame de Fourvière        & 92   & 318  \\
Notre-Dame de Paris            & 174  & 730  \\
Notre-Dame de Paris (Interior) & 299  & 976  \\
Notre-Dame de Strasbourg       & 122  & 437  \\
Opéra Garnier                  & 91   & 332  \\
Patio de los Arrayanes         & 107  & 328  \\
Patio de los Leones            & 126  & 306  \\
Pavillion de Flore (Louvre)    & 63   & 212  \\
Pont Alexandre III             & 55   & 131  \\
Pont des Arts                  & 74   & 157  \\
Saint-Martin, Colmar           & 161  & 692  \\
Salzburg Cathedral             & 98   & 303  \\
St. Mark's Basilica            & 79   & 328  \\
St. Paul's Cathedral           & 82   & 321  \\
Statue of Liberty              & 38   & 113  \\
Sukiennice                     & 55   & 186  \\
Taj Mahal                      & 68   & 228  \\
Torre de Belém                 & 125  & 457  \\
Umayyad Mosque (Courtyard)     & 76   & 252  \\
White House                    & 70   & 186 \\\bottomrule
\end{tabular}
\vspace{2pt}
\caption{Landmarks in the \ourdataset training set. We present the average and 95th percentile number of matches per scene. }
\label{tab:train_set}
\end{table}

\begin{table}[]
\begin{tabular}{lcc}
\toprule
Test scene                           & Mean & 95\%   \\ \midrule
Alexander Nevsky Cathedral, Lódz      & 47   & 115  \\
Alexander Nevsky Cathedral, Prešov    & 62   & 231  \\
Alexander Nevsky Cathedral, Sofia     & 87   & 244  \\
Alexander Nevsky Cathedral, Tallinn   & 53   & 162  \\
Arc de Triomphe de l'Étoile           & 100  & 387  \\
Berlin Cathedral                      & 77   & 372  \\
Brandenburg Gate                      & 36   & 95   \\
Cathedral of St. Peter and Paul, Brno & 283  & 1458 \\
Charlottenburg Palace                 & 40   & 104  \\
Church of the Saviour on the Blood    & 53   & 195  \\
Deutscher and Französischer Dom       & 67   & 139  \\
Florence Cathedral                    & 116  & 340  \\
Sleeping Beauty Castle                & 37   & 112  \\
St. Vitus Cathedral                   & 138  & 666  \\
Sydney Harbour Bridge                 & 29   & 104  \\
Washington Square Arch                & 70   & 206  \\ \bottomrule
\end{tabular}
\vspace{2pt}
\caption{Landmarks in the \ourdataset test set. We present the average and 95th percentile number of matches per scene. }
\label{tab:test_set}
\end{table}
\subsection{Implementation details of baselines}

We first provide additional details about the baselines evaluated in the main paper, then describe additional baselines provided in this supplemental material. 
We also evaluate two additional baselines, D2-Net~\cite{dusmanu2019d2}+RANSAC~\cite{fischler1981random} and SuperPoint~\cite{detone2018superpoint}+SuperGlue~\cite{sarlin2020superglue}, with results provided in Section~\ref{sec:quant_result}. 
With these two additional baselines, we cover a large variety of feature matching methods, 
including classical feature detectors such as SIFT and learning-based feature detectors such as D2-Net and SuperPoint. 
We also include traditional matching methods using nearest neighbor and RANSAC algorithms, as well as a learning-based matching method (SuperGlue). 
In addition, we evaluate detector-based feature matching methods and detector-free feature matching methods, such as LoFTR.
Note that all local feature matching baselines are used as classifiers on image pairs by thresholding either the number of matches, or the ratio of number of matches to number of keypoints.

\medskip \noindent \textbf{SIFT~\cite{lowe2004distinctive}+RANSAC~\cite{fischler1981random}.}
We use the COLMAP~\cite{schonberger2016structure} feature extraction and matching modules to produce keypoints and matches, using the default parameters. 
This includes the maximum extracted features set to 8192, use of cross check for matching, and geometric verification with a reprojection error of 4 and confidence level of 0.999.

\medskip \noindent \textbf{LoFTR~\cite{sun2021loftr}.} 
We follow the same process as previously described to use LoFTR to obtain matches for our network input.

\medskip \noindent \textbf{DINO~\cite{caron2021emerging}.} 
We use the pretrained ViT~\cite{dosovitskiy2020vit} small version model with a patch size of 16.
We pass one image at a time to DINO and obtain the latent code and feature maps from the last layer. 
We then train a linear classifier by taking the concatenated latent codes of images in a pair as input to a fully connected layer and outputting the probability.
For the feature maps, we concatenate them and pass them through a residual layer and fully connected layer to obtain the prediction.

\medskip \noindent \textbf{D2-Net~\cite{dusmanu2019d2}+RANSAC~\cite{fischler1981random}.}
D2-Net is a learning-based method for feature detector and descriptor.
We use its pretraine model on MegaDepth~\cite{li2018megadepth} to extract keypoints and descriptors.
We then use the OpenCV implementation of the brute force k-nearest neighbor matcher with cross-check setting, and apply a ratio test with a threshold of 0.75. 
Finally, we perform geometric verification with the same settings as previously described.

\medskip \noindent \textbf{SuperPoint~\cite{detone2018superpoint}+SuperGlue~\cite{sarlin2020superglue}.}
SuperPoint is an efficient learning-based method for detecting and describing keypoints.
Given the keypoints and descriptors extracted from SuperPoint, we use SuperGlue to obtain matches, where SuperGlue is a learning-based approach for feature matching using a graph neural network (GNN). 
We use the checkpoint trained for outdoor scenes with the recommended settings for SuperGlue, including a maximum number of keypoints set to 2048 and a Non-Maximum Suppression (NMS) radius of 3.

\begin{table*}[t]
\centering
\resizebox{\textwidth}{!}{
\begin{tabular}{lccccccccccc}
\toprule
\multirow{2}{*}{Average Precision}        & \multicolumn{2}{c}{D2-Net~\cite{dusmanu2019d2}+RANSAC~\cite{fischler1981random}} & \multicolumn{2}{c}{SuperPoint~\cite{detone2018superpoint}+SuperGlue~\cite{sarlin2020superglue}} & \multicolumn{2}{c}{SIFT~\cite{lowe2004distinctive}+RANSAC~\cite{fischler1981random}} & \multicolumn{2}{c}{LoFTR~\cite{sun2021loftr}} & \multicolumn{2}{c}{DINO~\cite{caron2021emerging}-ViT} & \multirow{2}{*}{Ours} \\
        & \#matches                                     & \%matches                                     & \#matches                                         & \%matches                                        & \#matches                                                       & \%matches                                                      & \#matches                        & \%matches                        & Latent code                          & Feature map                          &                       \\ \midrule
Average of all pairs from 16 landmarks     & 62.3                                       & 62.5                                       & 79.6                      & 80.7                      & 83.4                                                      & 81.2                                                           & 85.3                          & 86.0                          & 62.0                              & 63.3                              & \textbf{95.2}      \\ \midrule
Alexander Nevsky Cathedral, Łódź           & 62.1                                       & 63.7                                       & 83.7                      & 83.8                      & 72.7                                                      & 75.9                                                           & 80.7                          & 80.4                          & 50.9                              & 50.3                              & \textbf{89.5}      \\
Alexander Nevsky Cathedral, Sofia          & 63.6                                       & 63.7                                       & 80.6                      & 80.7                      & 89.5                                                      & 87.6                                                           & 90.0                          & 92.2                          & 53.0                              & 53.6                              & \textbf{98.5}      \\
Alexander Nevsky Cathedral, Tallinn        & 64.1                                       & 64.4                                       & 73.9                      & 74.3                      & 73.1                                                      & 76.0                                                           & 76.1                          & 80.3                          & 58.8                              & 50.8                              & \textbf{86.2}      \\
Arc de Triomphe                            & 49.7                                       & 48.8                                       & 60.1                      & 60.9                      & 86.1                                                      & 81.7                                                           & 85.7                          & 93.3                          & 55.4                              & 61.1                              & \textbf{97.6}      \\
Berlin Cathedral                           & 69.0                                       & 69.7                                       & 94.4                      & 94.6                      & 91.8                                                      & 91.6                                                           & 93.6                          & 92.7                          & 76.4                              & 70.6                              & \textbf{99.4}      \\
Brandenburg Gate                           & 42.1                                       & 42.8                                       & 60.1                      & 62.6                      & 79.3                                                      & 73.7                                                           & 90.9                          & 95.6                          & 60.8                              & 60.9                              & \textbf{99.8}      \\
Cathedral of Saints Peter and Paul in Brno & 75.0                                       & 74.6                                       & 93.1                      & 93.1                      & 95.8                                                      & 96.4                                                           & 89.8                          & 88.4                          & 64.6                              & 79.9                              & \textbf{99.8}      \\
Cathedral of St Alexander Nevsky, Prešov   & 73.2                                       & 74.5                                       & 89.8                      & 89.8                      & 82.5                                                      & 74.0                                                           & 86.1                          & 85.3                          & 62.9                              & 64.8                              & \textbf{94.6}      \\
Charlottenburg Palace                      & 62.0                                       & 60.6                                       & 81.4                      & 82.3                      & 81.5                                                      & 76.1                                                           & 85.6                          & 81.1                          & 65.8                              & 54.1                              & \textbf{93.3}      \\
Church of Savior on the Spilled Blood      & 62.1                                       & 61.0                                       & 86.7                      & 86.2                      & 82.1                                                      & 73.2                                                           & 84.9                          & 75.5                          & 63.9                              & 67.5                              & \textbf{93.8}      \\
Deutscher und Französischer Dom (Berlin)   & 53.9                                       & 54.6                                       & 75.4                      & 75.9                      & 74.5                                                      & 71.9                                                           & 85.8                          & 84.2                          & 55.6                              & 51.5                              & \textbf{98.1}      \\
Florence Cathedral                         & 60.1                                       & 58.0                                       & 82.7                      & 82.8                      & 90.6                                                      & 83.8                                                           & 84.5                          & 82.0                          & 54.6                              & 63.8                              & \textbf{94.2}      \\
Sleeping Beauty Castle                     & 54.4                                       & 56.8                                       & 71.0                      & 81.1                      & 81.1                                                      & 81.2                                                           & 75.0                          & 85.6                          & 67.2                              & 66.4                              & \textbf{97.1}      \\
St. Vitus Cathedral                        & 68.8                                       & 67.6                                       & 91.7                      & 91.0                      & 96.8                                                      & 88.0                                                           & 89.2                          & 87.5                          & 84.0                              & 77.0                              & \textbf{99.8}      \\
Sydney Harbour Bridge                      & 73.5                                       & 77.3                                       & 83.6                      & 86.8                      & 79.4                                                      & \textbf{92.3}                                                  & 83.8                          & 86.2                          & 53.0                              & 75.5                              & 87.0               \\
Washington Square Arch                     & 63.6                                       & 62.4                                       & 65.0                      & 65.5                      & 77.7                                                      & 75.9                                                           & 82.8                          & 86.0                          & 65.2                              & 65.0                              & \textbf{95.1}     

\\ \bottomrule
\end{tabular}
}
\vspace{2pt}
\caption{
Quantitative results for visual disambiguation evaluated on \ourdataset. Results are reported as the average precision (AP) multiplied by 100. We report both the average and the per-scene results for 16 landmarks.
}

\label{tab:full_quantitative_supp_ap}
\end{table*}

\begin{table*}[t]
\centering
\resizebox{\textwidth}{!}{
\begin{tabular}{lccccccccccc}
\toprule
\multirow{2}{*}{ROC AUC}         & \multicolumn{2}{c}{D2-Net~\cite{dusmanu2019d2}+RANSAC~\cite{fischler1981random}} & \multicolumn{2}{c}{SuperPoint~\cite{detone2018superpoint}+SuperGlue~\cite{sarlin2020superglue}} & \multicolumn{2}{c}{SIFT~\cite{lowe2004distinctive}+RANSAC~\cite{fischler1981random}} & \multicolumn{2}{c}{LoFTR~\cite{sun2021loftr}} & \multicolumn{2}{c}{DINO~\cite{caron2021emerging}-ViT} & \multirow{2}{*}{Ours} \\
        & \#matches                                     & \%matches                                     & \#matches                                         & \%matches                                        & \#matches                                                       & \%matches                                                      & \#matches                        & \%matches                        & Latent code                          & Feature map                          &                       \\ \midrule
Average of all pairs from 16 landmarks     & 53.5                                       & 53.7                                       & 76.8                       & 76.9                       & 80.2                                                      & 77.1                                                           & 78.9                          & 80.3                          & 60.9                              & 61.5                              & \textbf{93.8}      \\ \midrule
Alexander Nevsky Cathedral, Łódź           & 58.5                                       & 60.1                                       & 78.7                       & 78.7                       & 69.7                                                      & 72.7                                                           & 73.9                          & 74.8                          & 49.2                              & 49.7                              & \textbf{87.0}      \\
Alexander Nevsky Cathedral, Sofia          & 57.1                                       & 57.7                                       & 82.5                       & 82.5                       & 87.7                                                      & 84.3                                                           & 86.2                          & 89.1                          & 53.8                              & 49.3                              & \textbf{98.0}      \\
Alexander Nevsky Cathedral, Tallinn        & 59.2                                       & 60.0                                       & 71.8                       & 71.9                       & 68.0                                                      & 71.7                                                           & 71.9                          & 74.5                          & 60.8                              & 52.2                              & \textbf{84.2}      \\
Arc de Triomphe                            & 39.8                                       & 38.5                                       & 44.7                       & 44.7                       & 81.6                                                      & 75.3                                                           & 78.5                          & 88.9                          & 53.7                              & 57.1                              & \textbf{96.9}      \\
Berlin Cathedral                           & 54.8                                       & 56.1                                       & 92.1                       & 92.2                       & 89.2                                                      & 88.6                                                           & 89.4                          & 88.1                          & 71.6                              & 67.7                              & \textbf{99.3}      \\
Brandenburg Gate                           & 33.7                                       & 35.2                                       & 64.3                       & 64.5                       & 77.9                                                      & 71.9                                                           & 87.5                          & 93.4                          & 60.7                              & 60.4                              & \textbf{99.8}      \\
Cathedral of Saints Peter and Paul in Brno & 61.2                                       & 60.4                                       & 89.6                       & 89.6                       & 94.0                                                      & 95.0                                                           & 84.4                          & 82.8                          & 62.7                              & 75.8                              & \textbf{99.8}      \\
Cathedral of St Alexander Nevsky, Prešov   & 62.3                                       & 64.8                                       & 87.4                       & 87.4                       & 77.0                                                      & 63.9                                                           & 77.4                          & 77.6                          & 68.9                              & 60.6                              & \textbf{92.4}      \\
Charlottenburg Palace                      & 52.0                                       & 50.5                                       & 78.0                       & 78.2                       & 76.7                                                      & 70.7                                                           & 80.2                          & 77.1                          & 65.9                              & 53.6                              & \textbf{92.2}      \\
Church of Savior on the Spilled Blood      & 50.2                                       & 49.2                                       & 77.5                       & 77.3                       & 77.7                                                      & 68.4                                                           & 78.6                          & 70.4                          & 61.5                              & 64.0                              & \textbf{92.5}      \\
Deutscher und Französischer Dom (Berlin)   & 51.8                                       & 52.1                                       & 79.2                       & 79.2                       & 70.7                                                      & 68.2                                                           & 80.4                          & 78.7                          & 58.7                              & 49.2                              & \textbf{97.6}      \\
Florence Cathedral                         & 52.7                                       & 49.5                                       & 78.2                       & 78.2                       & 88.7                                                      & 80.3                                                           & 74.6                          & 71.9                          & 51.3                              & 62.5                              & \textbf{92.5}      \\
Sleeping Beauty Castle                     & 48.7                                       & 52.7                                       & 77.0                       & 77.5                       & 76.8                                                      & 79.4                                                           & 64.7                          & 78.4                          & 64.7                              & 66.5                              & \textbf{96.0}      \\
St. Vitus Cathedral                        & 49.6                                       & 47.4                                       & 82.4                       & 82.4                       & 96.7                                                      & 82.5                                                           & 82.3                          & 80.3                          & 80.4                              & 80.5                              & \textbf{99.8}      \\
Sydney Harbour Bridge                      & 69.4                                       & 71.8                                       & 86.3                       & 86.3                       & 76.3                                                      & \textbf{91.7}                                                  & 77.4                          & 79.0                          & 50.2                              & 72.4                              & 80.0               \\
Washington Square Arch                     & 55.9                                       & 53.9                                       & 59.6                       & 59.6                       & 73.9                                                      & 69.3                                                           & 74.7                          & 79.2                          & 59.9                              & 63.2                              & \textbf{93.5}     

\\ \bottomrule
\end{tabular}
}
\vspace{2pt}
\caption{
Quantitative results for visual disambiguation evaluated on \ourdataset. Results are reported as ROC AUC multiplied by 100. We report both the average and the per-scene results for 16 landmarks.
}
\label{tab:full_quantitative_supp_roc}
\end{table*}

\begin{table*}
\centering
\resizebox{\textwidth}{!}{
\begin{tabular}{lccccccccccccc}
\toprule
                                                        & \multirow{2}{*}{Images} & \multirow{2}{*}{COLMAP ~\cite{schonberger2016structure}} & \multirow{2}{*}{\cite{schonberger2016structure} \#matches$>$150} & \multirow{2}{*}{Heinly et al.~\cite{heinly2014correcting}} & \multirow{2}{*}{Wilson et al.~\cite{wilson2013network}} & \multirow{2}{*}{Cui et al.~\cite{cui2015global}} & \multirow{2}{*}{Yan et al.~\cite{yan2017distinguishing}} & \multicolumn{6}{c}{Ours}                                               \\ \cline{9-14}
                                                        &                         &                                                          &                                                                         &                                                            &                                                         &                                                  &                                                          & @0.5      & @0.6      & @0.7      & @0.8      & @0.9      & @0.97      \\ 
                                                        \midrule
Alexander Nevsky Cathedral~\cite{heinly2014correcting}  & 448                     & \ding{55}                                                & \ding{55}                                                               & \ding{51}                                                  & \ding{55}                                               & \ding{51}                                        & \ding{51}                                                & \ding{51} & \ding{51} & \ding{51} & \ding{51} & \ding{51} & \ding{51}  \\
Arc de Triomphe~\cite{heinly2014correcting}             & 434                     & \ding{55}                                                & \ding{55}                                                               & \ding{51}                                                  & \ding{55}                                               & \ding{55}                                        & \ding{51}                                                & \ding{51} & \ding{51} & \ding{51} & \ding{51} & \ding{51} & \ding{51}  \\
Berliner Dom~\cite{heinly2014correcting}                & 1,618                   & \ding{55}                                                & \ding{51}                                                               & \ding{51}                                                  & \ding{55}*                                              & \ding{51}                                        & \ding{55}*                                               & \ding{51} & \ding{51} & \ding{51} & \ding{51} & \ding{51} & \ding{55}* \\
Big Ben~\cite{heinly2014correcting}                     & 402                     & \ding{55}                                                & \ding{55}                                                               & \ding{51}                                                  & \ding{55}                                               & \ding{51}                                        & \ding{55}                                                & \ding{51} & \ding{51} & \ding{51} & \ding{51} & \ding{51} & \ding{51}  \\
Brandenburg Gate~\cite{heinly2014correcting}            & 175                     & \ding{55}                                                & \ding{51}                                                               & \ding{51}                                                  & --                                                      & \ding{55}                                        & \ding{55}                                                & \ding{55} & \ding{51} & \ding{51} & \ding{51} & \ding{51} & \ding{51}  \\
Church on the spilled blood~\cite{heinly2014correcting} & 277                     & \ding{55}                                                & \ding{55}                                                               & \ding{51}                                                  & --                                                      & \ding{55}                                        & \ding{55}                                                & \ding{55} & \ding{55} & \ding{55} & \ding{55} & \ding{55} & \ding{51}  \\
Radcliffe camera~\cite{heinly2014correcting}            & 282                     & \ding{55}                                                & \ding{51}                                                               & \ding{51}                                                  & \ding{55}*                                              & \ding{51}                                        & \ding{51}                                                & \ding{55} & \ding{51} & \ding{51} & \ding{51} & \ding{51} & \ding{51}  \\ \midrule
Number of scenes: \ding{51}/\ding{55}*/\ding{55}        &                         & 0/0/7                                                    & 3/0/4                                                                   & 7/0/0                                                      & 0/2/3                                                   & 4/0/3                                            & 3/1/3                                                    & 4/0/3     & 6/0/1     & 6/0/1     & 6/0/1     & 6/0/1     & 6/1/0      
\\ \bottomrule
\end{tabular}
}
\vspace{2pt}
\caption{
Robustness evaluation of our method to the probability threshold on SfM disambiguation results.
\ding{51} means correctly disambiguate and reconstruct. \ding{55} means fail to disambiguate and \ding{55}* means over-split.
Our method exhibits robustness to the probability threshold and successfully reconstructed 6 out of 7 scenes with probability thresholds ranging from 0.6 to 0.97.
}
\label{tab:sfm_comparison_supp}
\end{table*}

\begin{figure}
\begin{center}
\includegraphics[width=0.95\columnwidth, trim=0 0 0 0, clip]{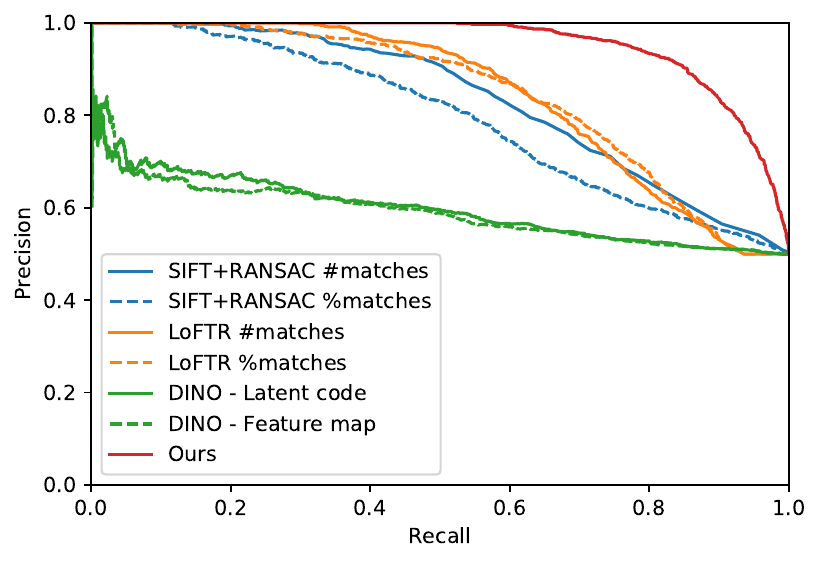}
\end{center}
\vspace{-10pt}
   \caption{
   Precision-Recall (PR) curves on the \ourdataset test set. 
   The $x$-axis represents recall and the $y$-axis represents precision. 
   A curve approaching the top-right corner indicates better performance.
   }
\label{fig:prcurve_supp}

\begin{center}
\includegraphics[width=0.95\columnwidth, trim=0 0 0 0, clip]{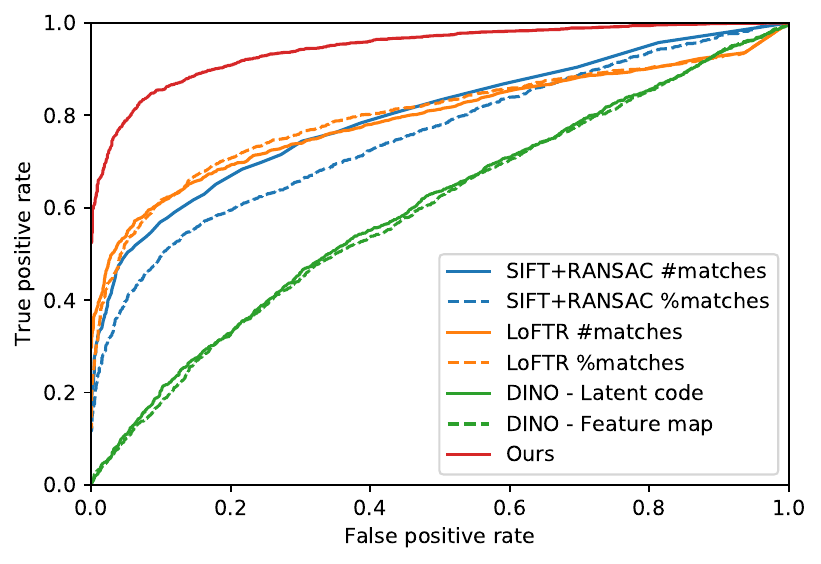}
\end{center}
\vspace{-10pt}
   \caption{
   Receiver operating characteristic (ROC) curves on the \ourdataset test set. 
   The $x$-axis represents the false positive rate, and the $y$-axis represents the true positive rate.
   The ideal method would simultaneously have a lower false positive rate and higher true positive rate, with the curve approaching the top-left corner. 
   }
\label{fig:roc_supp}
\end{figure}

\begin{figure}
\begin{center}
\includegraphics[width=0.9\columnwidth, trim=0 0 0 0, clip]{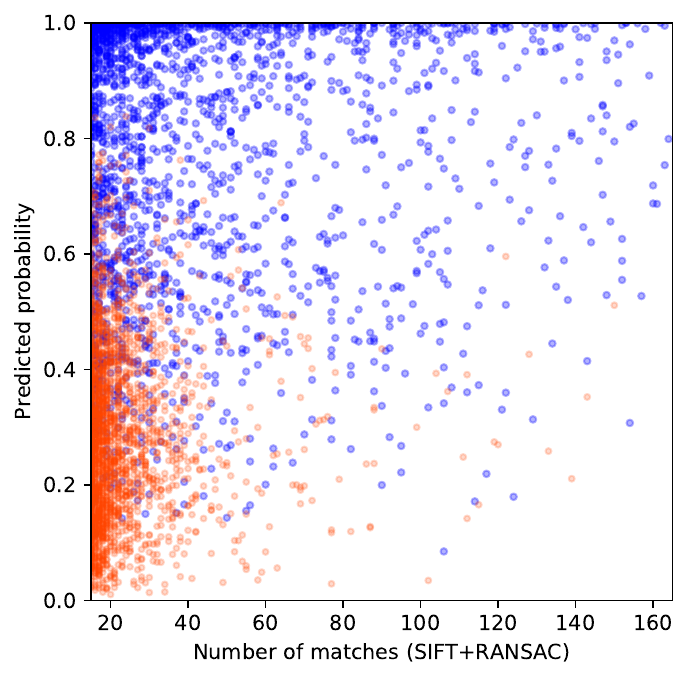}
\includegraphics[width=0.9\columnwidth, trim=0 0 0 0, clip]{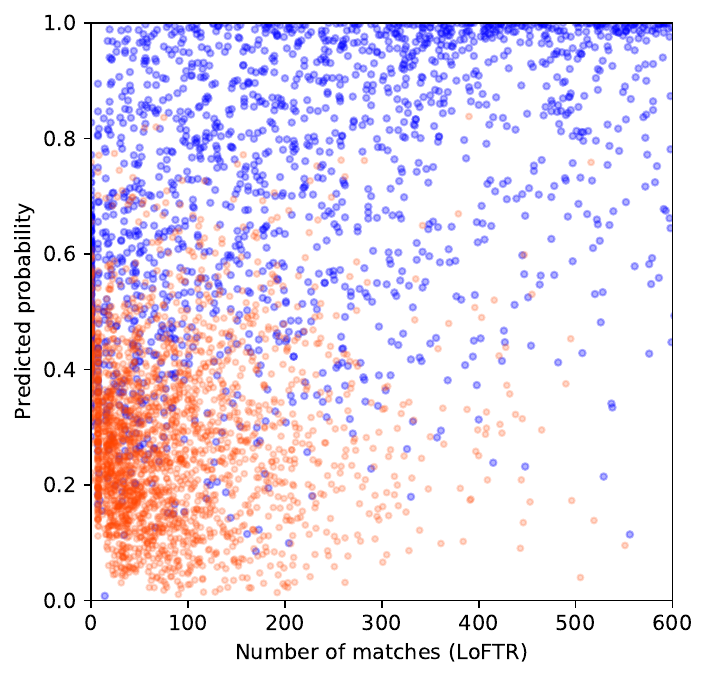}
\end{center}
\vspace{-10pt}
   \caption{
   Correlation between predicted probability of our network and number of matches using SIFT+RANSAC (top figure) and LoFTR (bottom figure) for pairs in all test scenes.
   The $x$-axis represents the number of matches and the $y$-axis represents the predicted probability.
   Blue dots represent ground truth positive pairs and red dots represent ground truth negative pairs. 
   Our method produces a high probability for most positive pairs and a low probability for most negative pairs. 
   Even for challenging doppelganger pairs, our method can produce a probability of less than 0.85.
   In contrast, SIFT and LoFTR methods have lots of positive pairs with low numbers of matches, as well as a number of negative pairs with large numbers of matches.
   }
\label{fig:correlation_supp}
\end{figure}

\begin{figure}
\begin{center}
\includegraphics[width=0.9\columnwidth, trim=0 0 0 0, clip]{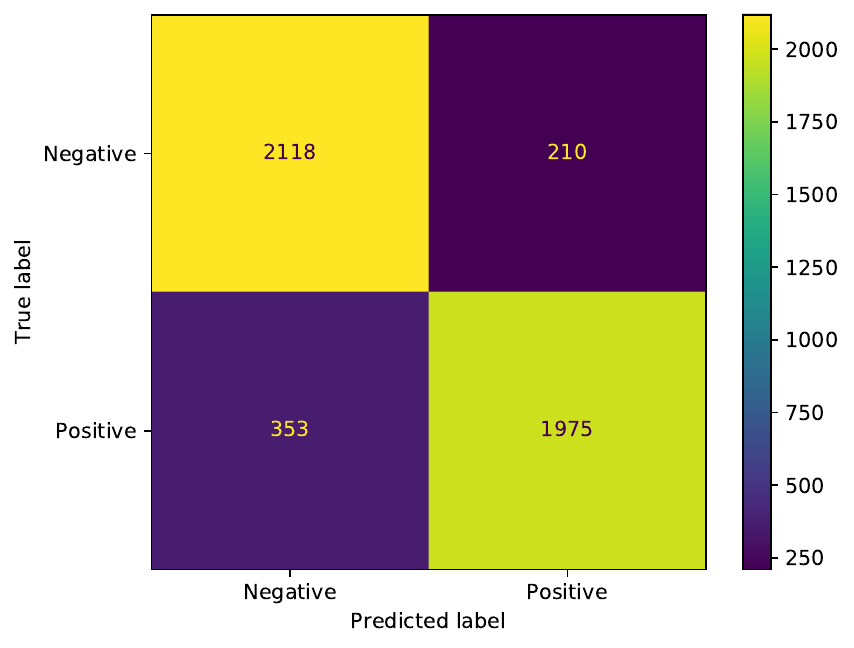}
\end{center}
\vspace{-10pt}
   \caption{
Confusion matrix for our method with probability threshold set to 0.5.
Our method correctly identifies 1,975 true positive pairs and 2,118 true negatives, 
while producing 353 false positive pairs and 210 false negatives.
Overall, the method achieves an accuracy of approximately 88\%.
   }
\label{fig:confusion_supp}
\end{figure}

\subsection{Quantitative results and analysis}
\label{sec:quant_result}

We present additional comparisons of our method with baselines on the \ourdataset test set, and report the average precision (AP) and ROC AUC scores in Tables~\ref{tab:full_quantitative_supp_ap} and \ref{tab:full_quantitative_supp_roc}, respectively.
Both the AP and ROC AUC scores evaluate classification performance, and higher scores are better.
While AP is more focused on positive pairs, ROC AUC is more focused on the ranking of predictions and cares equally about positive and negative pairs.
Our method outperforms all other baselines for 15 out of 16 test landmarks, with an average precision (AP) of 95.2\% and an ROC AUC of 93.8\% across all landmarks.
The SuperPoint+SuperGlue method achieves comparable results to SIFT+RANSAC, while D2-Net performs worse and similarly to DINO. 
These results suggest that the presence of the number or ratio of matches is not necessarily the best indicator of whether two images truly match.

We also present evaluation results as the precision-recall (PR) curves shown in Figure~\ref{fig:prcurve_supp}, where our method shows significant improvements over the baselines.
We also provide receiver operating characteristic (ROC) curves in Figure~\ref{fig:roc_supp}. 
ROC curves illustrate the performance of classifiers across various classification thresholds. 
Our model consistently outperforms other methods across all thresholds, with the lowest false positive rate and highest true positive rate.
Additionally, in Figure~\ref{fig:confusion_supp}, we show the confusion matrix of our network predictions using a threshold of 0.5, indicating that our method can correctly classify approximately 88\% of image pairs in the test set at this threshold.

To analyze the correlation between our network's predictions and the number of matches in the input pair,
we generate 2D scatter plots where the $x$-axis is the number of matches and the $y$-axis is the probability predicted by our network.
The resulting scatter plots are shown in Figure~\ref{fig:correlation_supp}, using SIFT+RANSAC and LoFTR methods to compute matches, respectively.
In the figure, red dots represent pairs with a ground truth label of negative, while blue dots represent positive pairs.
The figure shows that our method can differentiate between positive and negative image pairs, in particular in cases when such pairs have the same number of matches.
Although differentiating doppelganger pairs with larger numbers of matches can be more challenging (red dots at top right of figure), our method still predicts a probability lower than 0.8 for most negative pairs.

\begin{table}[]
\centering
\begin{tabular}{lr}
\toprule
Full                  & 95.2 \\ \midrule
w/o Augmentation      & 93.6 \\
w/o Masks              & 64.7 \\
w/o RGB               & 90.0 \\
w/o Geo. verification & 92.1 \\ \bottomrule
\end{tabular}
\vspace{2pt}
\caption{Additional ablation study on network input design. The results are reported as the average precision multiplied by 100.}
\label{tab:ablation_supp}
\end{table}

\subsection{Additional ablation study}
We conduct an additional ablation study on the design of network input. 
The results, reported as average precision scores, are shown in Table \ref{tab:ablation_supp}.
As described in the main paper, we conduct a \textit{w/o Augmentation} experiment where we train the classifier on 44 scenes without flip augmentations.
The remaining variations are trained on the same dataset of 44 scenes without augmentation for speed of training.

In the \textit{w/o Masks} setting, we remove keypoint and match masks from input, leaving only RGB images. 
This results in significant degeneration of performance.
In the \textit{w/o RGB} experiment, we remove RGB images from the input, leaving only keypoint and match masks.
This leads to a drop in average precision from 93.6\% to 90.0\%. This drop is not as significant as that stemming from removal of keypoint and match masks, indicating the relative importance of these inputs.
The \textit{w/o Geo.\ verification} setting is one where matches are not filtered and verified with Fundamental matrix estimation using RANSAC, resulting in a decrease in average prevision from 93.6\% to 92.1\%.
In summary, the ablation study demonstrates that keypoint and match masks are essential components of input for visual disambiguation, as they contain rich information and cues for differentiating visually similar pairs.

\subsection{Additional qualitative results}
\begin{figure*}
\begin{minipage}[b]{.495\textwidth}
\begin{center}
\includegraphics[width=\columnwidth, trim=0 0 0 0, clip]{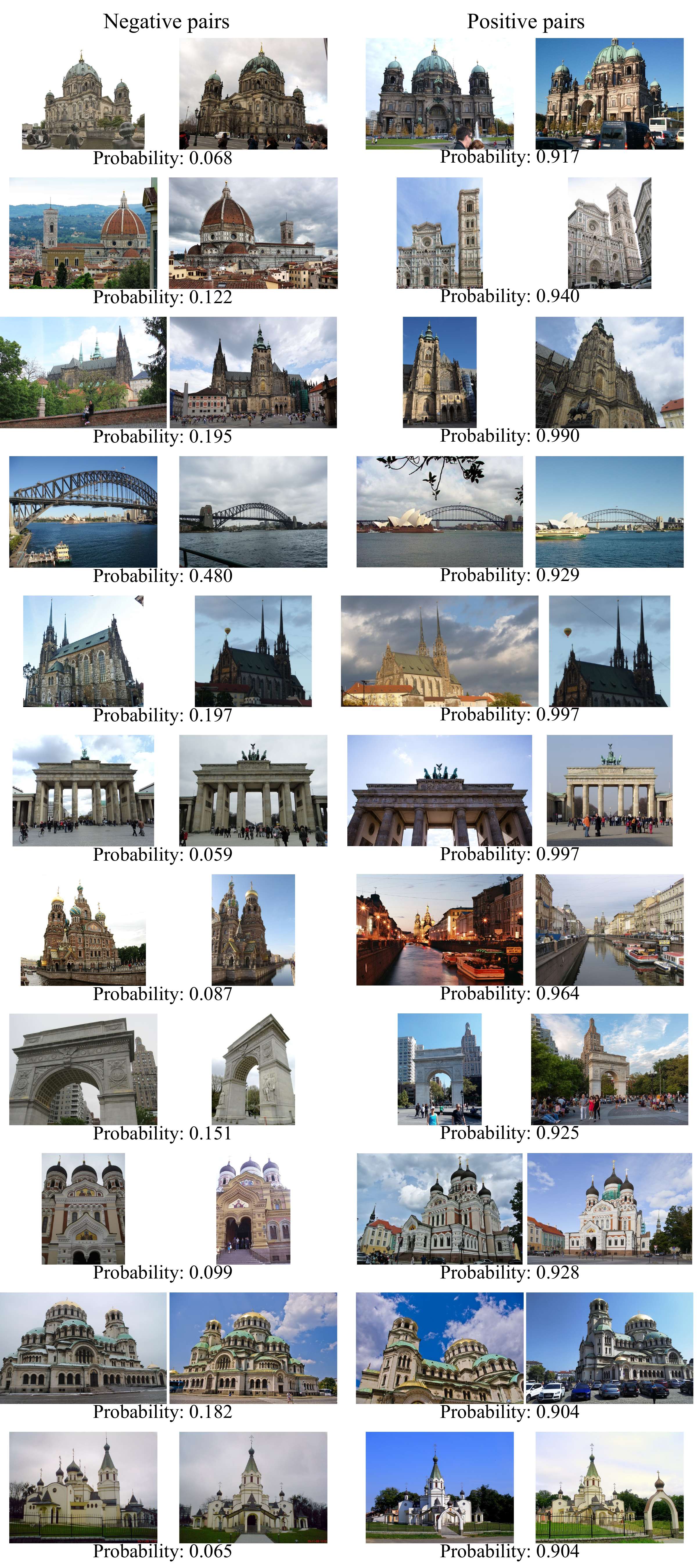}
\end{center}
   \caption{
   Additional visual disambiguation results. We visualize test image pairs with their corresponding predicted probabilities produced by our network. The left column shows negative pairs and the right column shows positive pairs.
   }
\label{fig:test_pair_supp}
\end{minipage}\qquad
\begin{minipage}[b]{.495\textwidth}
\begin{center}
\includegraphics[width=0.85\columnwidth, trim=0 0 0 0, clip]{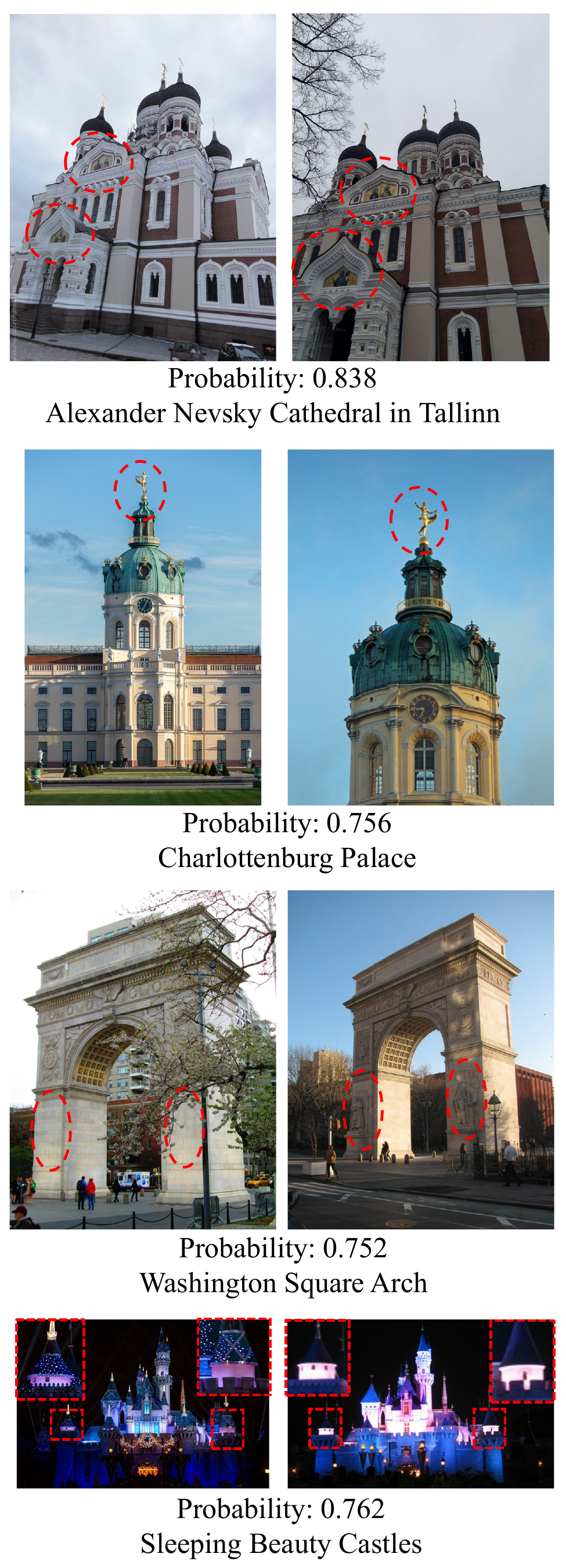}
\end{center}
\vspace{-10pt}
   \caption{
   Failure cases. 
   We visualize challenging \twinterm pairs that are all negative pairs, but the predicted probabilities by our network are high. 
   We circle the regions that might be helpful for disambiguation in red. 
   For the last pair from Sleeping Beauty Castles, we show zoomed-in views of distinct regions with red boxes.
   }
\label{fig:fail_case_supp}
\end{minipage}
\end{figure*}

In Figure~\ref{fig:test_pair_supp}, we provide additional visualizations of test image pairs and their corresponding predicted probability by our method on a variety of test scenes.

We visualize some failure cases in Figure~\ref{fig:fail_case_supp}, all of which are negative pairs.
We circle potentially useful regions for visual disambiguation in red. 
The pair from Alexander Nevsky Cathedral in Tallinn has distinct regions on the facades that are difficult to observe due to the viewpoint. 
Given other regions and structures of the building appear similar, it is challenging even for humans to differentiate between the images.
In the pair from Charlottenburg Palace, the second image is a zoom-in view that crops out other regions, leaving only a small region on the golden sculpture (at the top of the building) that can serve as a cue for visual disambiguation.
In the third pair from Washington Square Arch, the illumination differences might mask the structural differences (which are in shadow in the second image), making it more difficult to discern the differences between regions.
The replicas of Sleeping Beauty Castles look very similar, as shown in the last pair of images.  
Images captured at night can be more challenging to distinguish, since the background is obscured and important cues may be lost due to lack of observability in the background.

\section{Structure from Motion disambiguation}

\subsection{Threshold robustness evaluation}
We evaluate the robustness of our method for disambiguating SfM reconstructions to the probability threshold, and we show additional results on 7 landmark datasets from Heinly et al.~\cite{heinly2014correcting} with thresholds at [0.5, 0.6, 0.7, 0.8, 0.9, 0.97] in Table~\ref{tab:sfm_comparison_supp}.
At the threshold of 0.5, some incorrect pairs are included in the scene graph, resulting in broken reconstructions for Brandenburg Gate, Church on Spilled Blood, and Radcliffe Camera.
For the SfM disambiguation setting, where a single bad matching pair can break a model, we care more about false positives than keeping all positive pairs (i.e., we care more about precision than recall). Therefore setting the threshold to 0.5 may intuitively not be the best strategy, hence the better performance at higher thresholds that filter out more pairs.
For thresholds ranging from 0.6 to 0.9, our method is robust, and successfully disambiguates and reconstructs 6 out of 7 scenes.
At even higher thresholds, we see that one of the models (Berliner Dom) splits apart, resulting in over-splitting of the reconstruction, 
but at this strict threshold we can successfully disambiguate the final scene (Church on Spilled Blood).
Overall, our method is able to reconstruct 6 out of 7 scenes even at this threshold, demonstrating the robustness and effectiveness of our approach.

\subsection{Detailed reconstruction visualization}
\begin{figure*}
\begin{minipage}{0.65\linewidth}
\begin{center}
\includegraphics[width=\columnwidth, trim=0 0 0 0, clip]{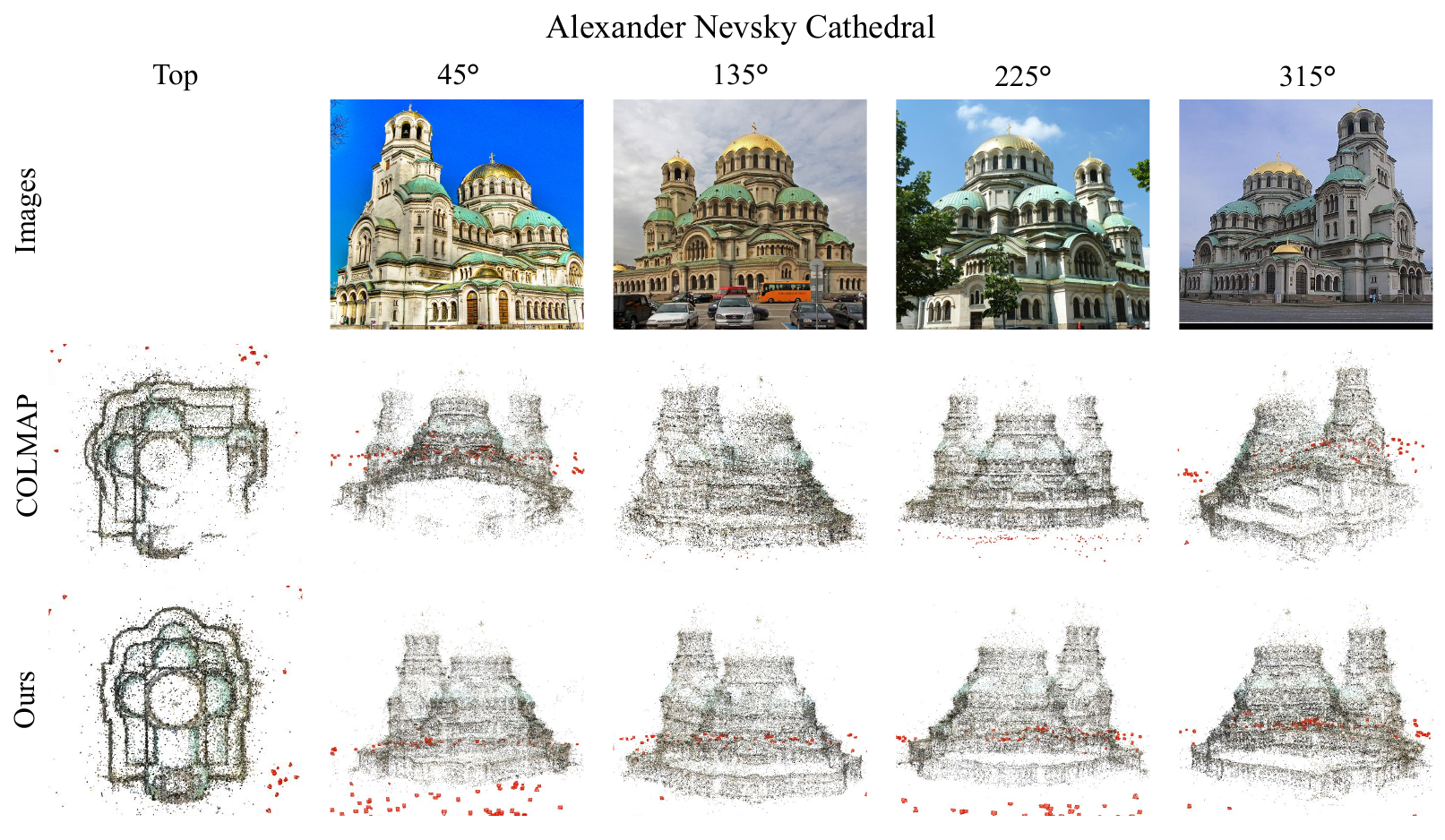}
\includegraphics[width=\columnwidth, trim=0 0 0 0, clip]{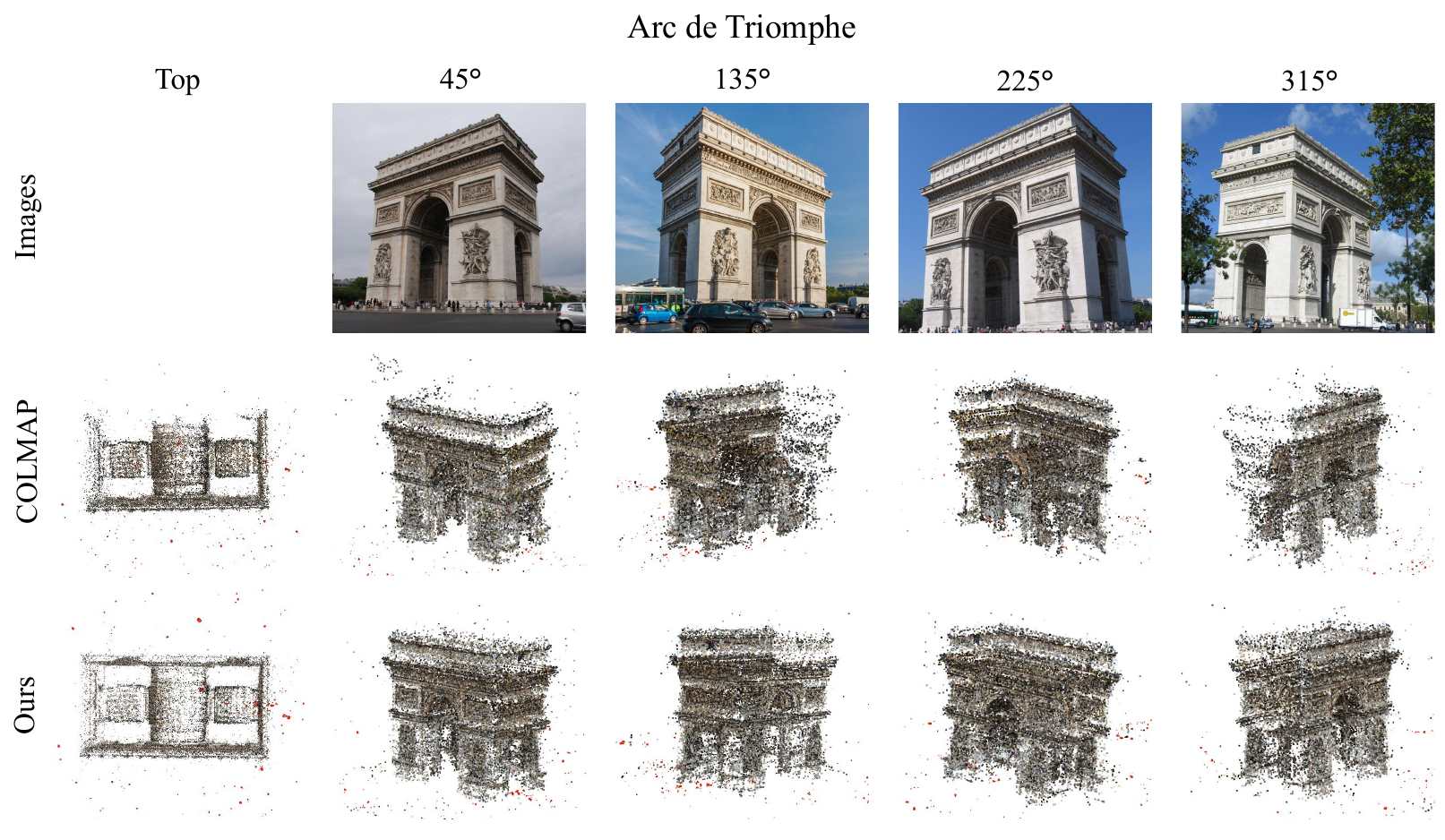}
\includegraphics[width=\columnwidth, trim=0 0 0 0, clip]{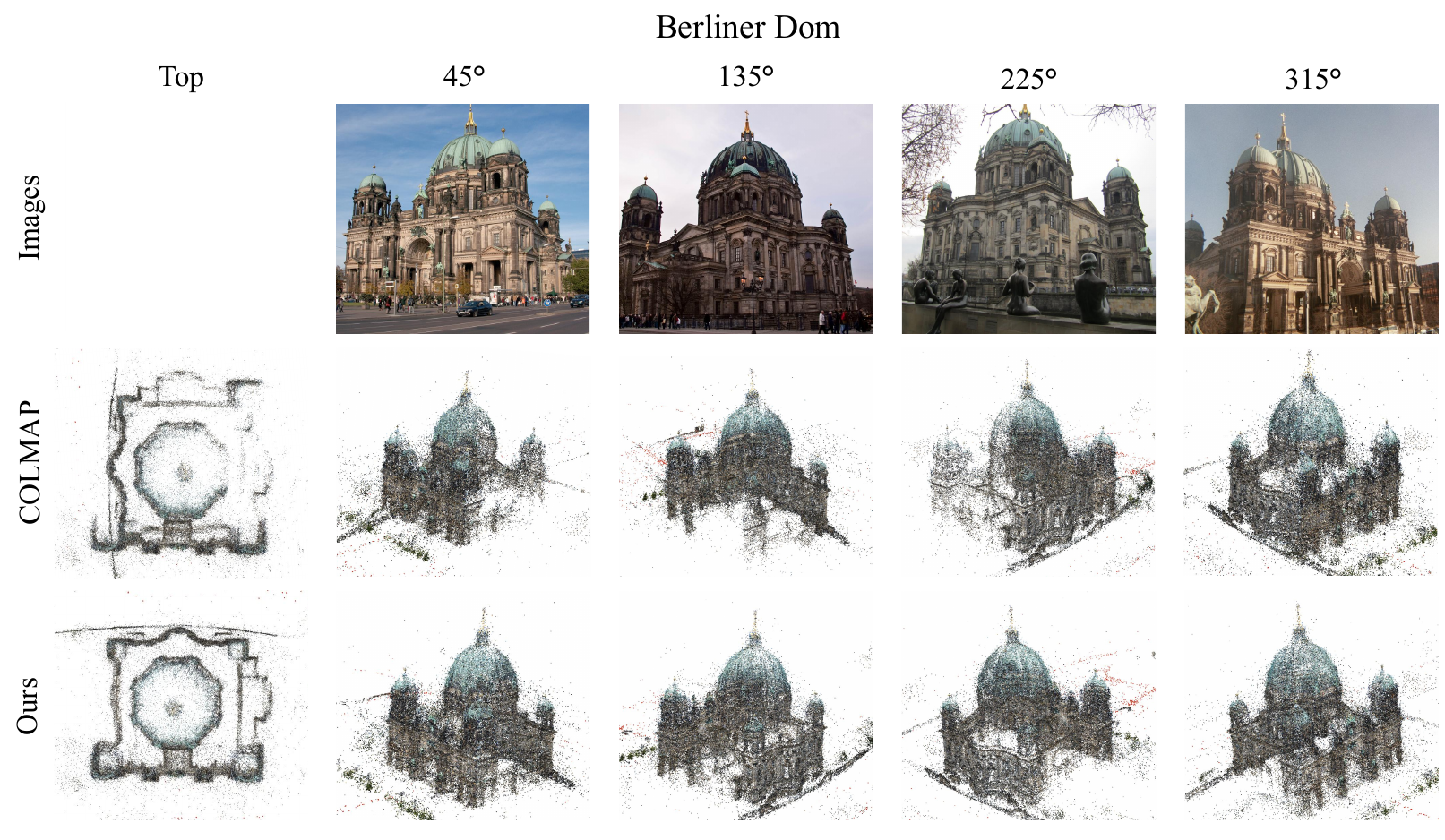}
\end{center}
\end{minipage}
\begin{minipage}{0.35\linewidth}
\begin{center}
\includegraphics[width=0.85\columnwidth, trim=0 0 0 0, clip]{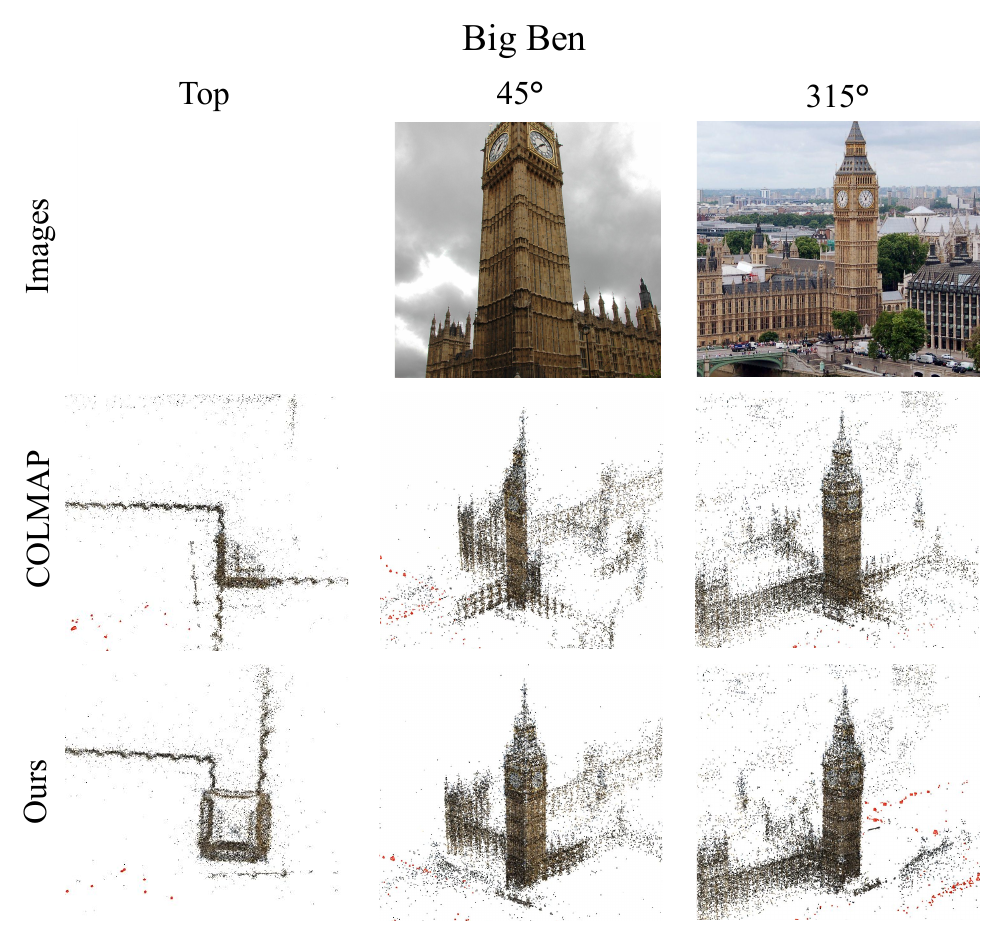}
\includegraphics[width=0.85\columnwidth, trim=0 0 0 0, clip]{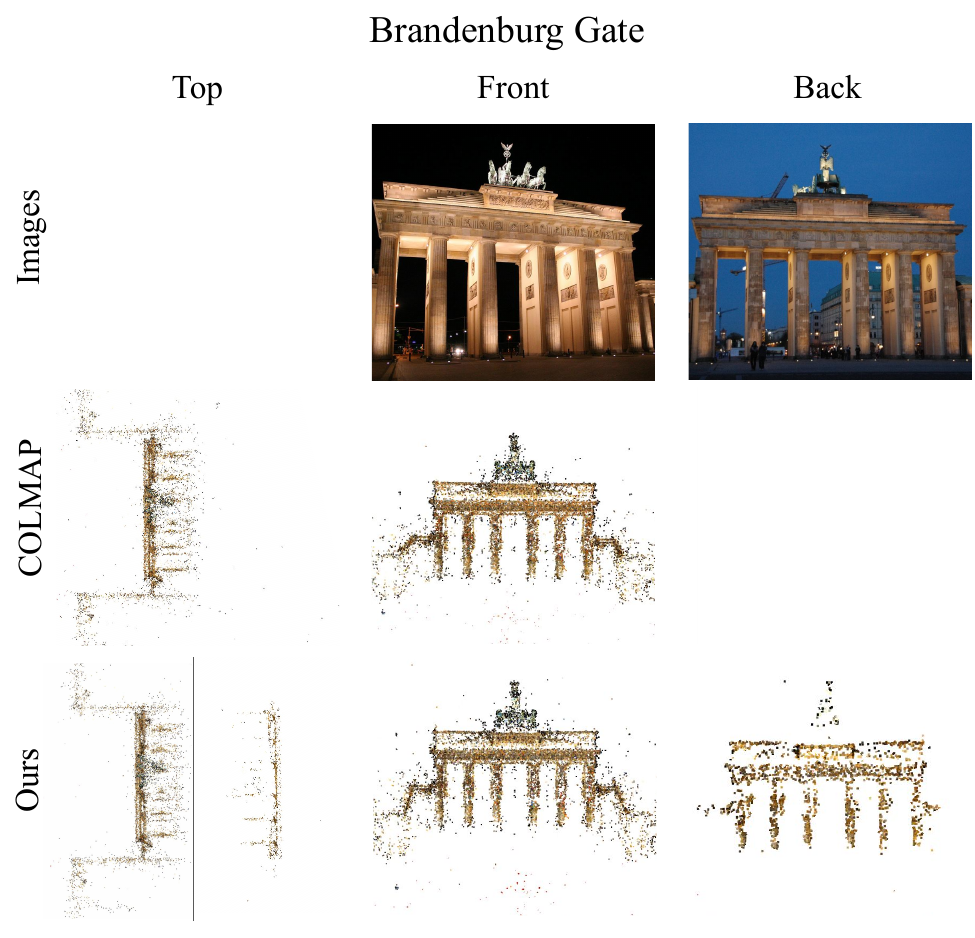}
\includegraphics[width=0.85\columnwidth, trim=0 0 0 0, clip]{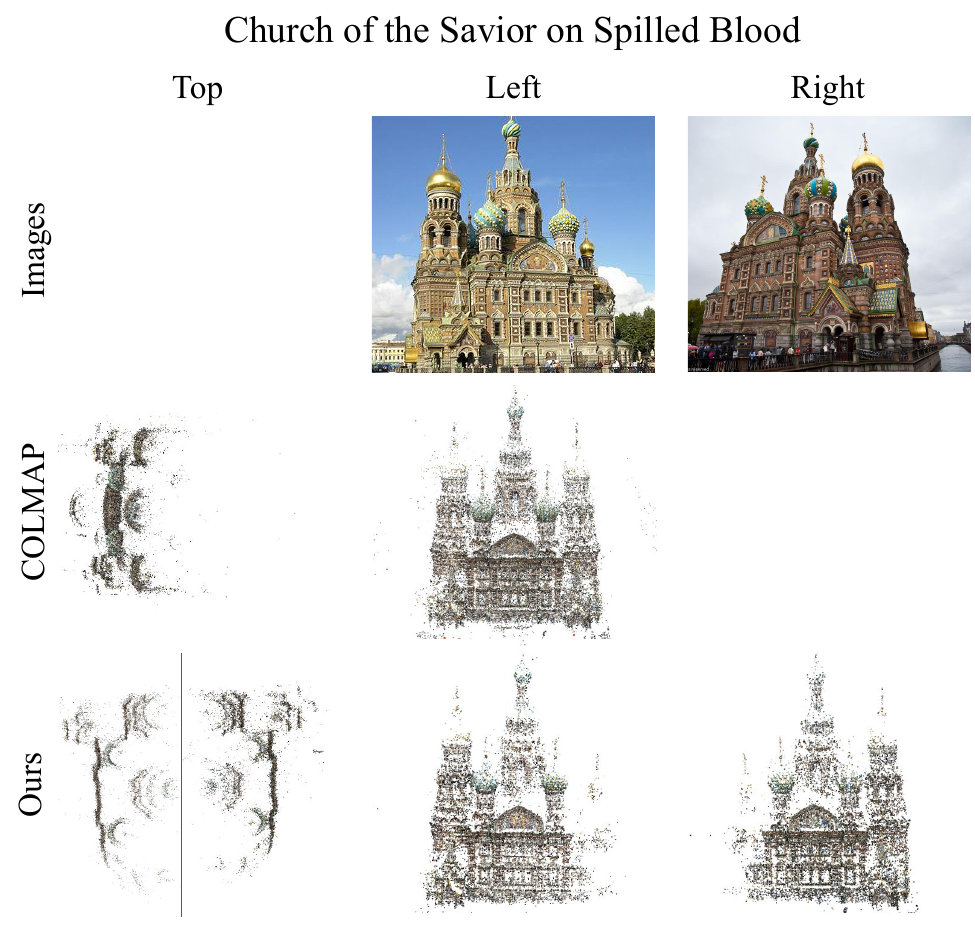}
\includegraphics[width=0.85\columnwidth, trim=0 0 0 0, clip]{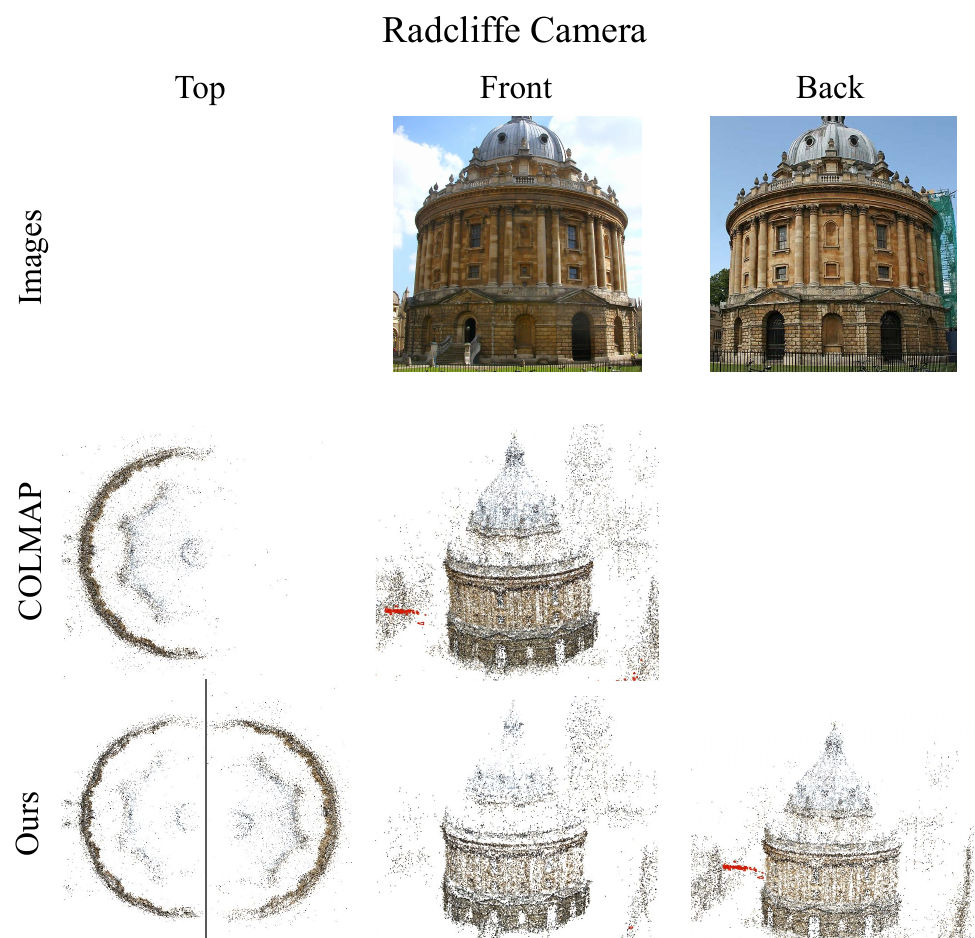}
\end{center}
\end{minipage}
   \caption{
   \small
   Visualization of Structure from Motion (SfM) disambiguation results from different viewpoints.
   We show a set of input RGB images at the top of each example scene,
   vanilla COLMAP reconstructions in the middle,
   and our method's disambiguated reconstructions at the bottom.
   For reconstructions where an angle is denoted, the 0\textdegree \xspace mark begins at the bottom of the birds-eye view and increases counterclockwise about the center of the image.
   Note that for some landmarks, the correct reconstruction is separated into two components when disambiguated due to a lack of camera views from sufficient viewpoints.
   }
\label{fig:colmap7scenes_supp}
\end{figure*}

We present a detailed visualization of the reconstruction results for 7 scenes rendered from different viewpoints in Figure~\ref{fig:colmap7scenes_supp}, comparing our method with vanilla COLMAP reconstruction.
The visualizations from different viewpoints provide a clear view of the incorrect structures produced by COLMAP, 
such as the double towers in the Alexander Nevsky Cathedral and the missing sides of Big Ben.
Our method can disambiguate different sides of these highly symmetric landmarks and produce a complete and correct reconstruction.

\end{document}